\documentclass{templates/apple/applemlr}
\usepackage{amsmath}
\usepackage{enumerate}
\usepackage{algorithm}
\usepackage{algpseudocode}
\usepackage{amsfonts}
\usepackage{amsthm}
\usepackage{newtxtt}
\usepackage{cleveref}
\usepackage{diagbox}
\usepackage{colortbl}
\usepackage{amssymb}
\usepackage{xspace}
\usepackage{wrapfig}
\usepackage{adjustbox}
\usepackage{tabularx}
\usepackage{booktabs}
\usepackage{mathtools}
\usepackage{tikz}
\usepackage{silence}
\usepackage{dsfont}
\usepackage[table]{xcolor}
\usepackage[dvipsnames]{xcolor}
\usepackage{multirow}
\usepackage{makecell}
\usepackage{xfakebold}

\definecolor{textgray}{HTML}{6E6E73}
\usetikzlibrary{positioning, calc}
\usetikzlibrary{decorations.pathmorphing}

\makeatletter
\patchcmd{\wrong@fontshape}{\@gobbletwo}{}{}{}
\makeatother
\numberwithin{equation}{section}
\makeatletter
\AtBeginDocument{
  \urlstyle{sf}
  
}
\makeatother

\definecolor{light}{RGB}{125, 125, 125}
\crefname{tcb@cnt@pbox}{code}{code}
\Crefname{tcb@cnt@pbox}{Code}{Code}
\crefname{assumption}{assumption}{assumption}
\Crefname{assumption}{Assumption}{Assumptions}

\newtcolorbox[auto counter]{pbox}[2][]{
  colback=white,
  title=Code~\thetcbcounter: #2,
  #1,fonttitle=\sffamily,
  fontupper=\sffamily,
  arc=2pt,
  colframe=bgcolor,
  coltitle=fgcolor,
  colbacktitle=bgcolor,
  toptitle=0.25cm,
  bottomtitle=0.125cm
}

\makeatletter
\newcommand\applefootnote[1]{%
  \begingroup
  \renewcommand\thefootnote{}%
  \renewcommand\@makefntext[1]{\noindent##1}%
  \footnote{#1}%
  \addtocounter{footnote}{-1}%
  \endgroup
}
\makeatother

\definecolor{cverbbg}{gray}{0.90}

\usepackage{amsmath, amsthm, amssymb}

\usepackage{titletoc}
\usepackage[page,header]{appendix}

\usepackage{expl3}  %
\ExplSyntaxOn
  _to_str:N
\ExplSyntaxOff
\usepackage{ifthen}
\usepackage{pgffor}

\usepackage[acronym,nowarn,section,nonumberlist]{glossaries}
\glsdisablehyper  %
\makeglossaries

\setacronymstyle{long-short}

\newacronym{auc}{AUC}{Area Under the Curve}
\newacronym{asr}{ASR}{automatic speech recognition}

\newacronym{bert}{BERT}{Bidirectional Encoder Representations from Transformers}
\newacronym{bow}{BOW}{Bag of Words}
\newacronym{boe}{BOE}{Bag of Embeddings}
\newacronym{borep}{BOREP}{Bag of Random Embedding Projections}
\newacronym{bn}{BN}{Bayesian Network}

\newacronym{cce}{CCE}{Cut Cross-Entropy}
\newacronym{cdf}{CDF}{Cumulative Distribution Function}
\newacronym{cnn}{CNN}{Convolutional Neural Network}
\newacronym{cka}{CKA}{Centered Kernel Alignment}
\newacronym{cvae}{CVAE}{Conditional Variational Auto-Encoder}

\newacronym{dag}{DAG}{Directed Acyclic Graph}
\newacronym{dgm}{DGM}{Directed Graphical Model}
\newacronym{dnn}{DNN}{Deep Neural Network}
\newacronym{dac}{DAC}{Descript Audio Codec}

\newacronym{ecfp4}{ECFP4}{Extended Connectivity Fingerprint}
\newacronym{ehr}{EHR}{Electronic Health Record}
\newacronym{esn}{ESN}{Echo State Network}
\newacronym{elbo}{ELBO}{variational evidence lower bound}

\newacronym{fft}{FFT}{Fast Fourier Transform}
\newacronym{fs}{FS}{FastSent}
\newacronym{ft}{FT}{Fourier Transform}

\newacronym{gat}{GAT}{Graph Attention Network}
\newacronym{gcn}{GCN}{Graph Convolutional Network}
\newacronym{gdl}{GDL}{Geometric Deep Learning}
\newacronym{ggnn}{GGNN}{Gated Graph Neural Network}
\newacronym{glove}{GloVe}{Global Vectors for Word Representation}
\newacronym{gnn}{GNN}{Graph Neural Network}
\newacronym{gru}{GRU}{Gated Recurrent Unit}

\newacronym{hmm}{HMM}{Hidden Markov Model}

\newacronym{idf}{IDF}{Inverse Document Frequency}
\newacronym{ig}{IG}{Information Gain}
\newacronym{isan}{ISAN}{Input Switched Affine Network}

\newacronym{jsd}{JSD}{Jensen-Shannon Divergence}

\newacronym{kld}{KLD}{Kullback-Leibler Divergence}
\newacronym{knn}{KNN}{$K$-Nearest Neighbour}

\newacronym{lhs}{L.H.S.}{left hand side}
\newacronym{lstm}{LSTM}{Long Short Term Memory Network}

\newacronym{map}{MAP}{Maximum a Posteriori}
\newacronym{mc}{MC}{Monte Carlo}
\newacronym{mdl}{MDL}{Minimum Description Length}
\newacronym{mdm}{MDM}{Masked Diffusion Models}
\newacronym{ml}{ML}{Machine Learning}
\newacronym{mlp}{MLP}{Multi-Layer Perceptron}
\newacronym{mle}{MLE}{Maximum Likelihood Estimation}
\newacronym{mnist}{MNIST}{Modified National Institute of Standards and Technology}
\newacronym{moe}{MoE}{Mixture of Experts}

\newacronym{nlp}{NLP}{Natural Language Processing}
\newacronym{nn}{NN}{Neural Network}
\newacronym{nll}{NLL}{Negative Log Likelihood}

\newacronym{ode}{ODE}{Ordinary Differential Equation}
\newacronym{oh}{OH}{One-Hot}

\newacronym{pdf}{PDF}{Probability Density Function}
\newacronym{pgm}{PGM}{Probabilistic Graphical Model}

\newacronym{relu}{ReLU}{Rectified Linear Unit}
\newacronym{rdf}{RDF}{Resource Description Framework}
\newacronym{rgb}{RGB}{Red Green Blue}
\newacronym{rgc}{RGC}{Relational Graph Convolution}
\newacronym{rgat}{RGAT}{Relational Graph Attention Network}
\newacronym{rgcn}{RGCN}{Relational Graph Convolutional Network}
\newacronym{rhs}{R.H.S.}{Right Hand Side}
\newacronym{rnn}{RNN}{Recurrent Neural Network}
\newacronym{roc}{ROC}{Receiver Operating Characteristic}
\newacronym{rvq}{RVQ}{residual vector quantized}

\newacronym{sgd}{SGD}{Stochastic Gradient Descent}
\newacronym{sg}{SG}{SkipGram}
\newacronym{simclr}{SimCLR}{Simple framework for Contrastive Learning of visual Representations}
\newacronym{st}{ST}{SkipThought}
\newacronym{sts}{STS}{Semantic Textual Similarity}
\newacronym{sde}{SDE}{stochastic differential equation}

\newacronym{termfreq}{TF}{Term Frequency}
\newacronym{tfidf}{TF-IDF}{Term Frequency-Inverse Document Frequency}
\newacronym[plural=TPPs, descriptionplural={Temporal Point Processes}]{tpp}{TPP}{Temporal Point Process}
\newacronym{tts}{TTS}{text-to-speech}

\newacronym{vae}{VAE}{Variational Auto-Encoder}

\newacronym{wirgat}{WIRGAT}{Within-Relation Graph Attention}
\newacronym{wl}{WL}{Weisfeiler-Lehman}

\usepackage{bm}
\usepackage[T1]{fontenc}  %
\usepackage[type1]{libertine}
\usepackage[libertine]{newtxmath}
\renewcommand{\sfdefault}{sfpro}
\usepackage{mathtools}

\usepackage{hyperref}
\usepackage{cleveref}

\usepackage{makecell}

\usepackage{paralist}

\usepackage{placeins}
\usepackage{titletoc}
\usepackage[page,header]{appendix}

\usepackage{subcaption}

\usepackage{booktabs}

\usepackage{alphalph}
\usepackage{multirow}

\DeclareRobustCommand{\mathup}[1]{\begingroup\changegreek\mathrm{#1}\endgroup}
\DeclareRobustCommand{\mathbfup}[1]{\begingroup\changegreekbf\mathbf{#1}\endgroup}
\DeclareRobustCommand{\mathbit}[1]{\bm{\mathit{#1}}}

\DeclareMathAlphabet{\mathsfit}{\encodingdefault}{\sfdefault}{m}{sl}
\SetMathAlphabet{\mathsfit}{bold}{\encodingdefault}{\sfdefault}{bx}{n}
\newcommand{\tens}[1]{\bm{\mathsfit{#1}}}

\newcommand{\constantvector}{\bm}               %

\newcommand{\constantmatrix}{\bm}               %
\newcommand{\constantmatrixgreek}{\mathbit}

\newcommand{\randomscalar}{\textnormal}         %
\newcommand{\randomscalargreek}{\mathup}

\newcommand{\randomvector}{\mathbf}             %
\newcommand{\randomvectorgreek}{\mathbfup}

\newcommand{\randommatrix}{\mathbf}             %
\newcommand{\randommatrixgreek}{\mathbfup}

\newcommand{\graphstyle}{\mathcal}              %

\newcommand{\tensorstyle}{\tens}                %

\newcommand{\setstyle}{\mathbb}                %

\usepackage{tcolorbox}
\tcbuselibrary{breakable}

\usepackage[most]{tcolorbox} %

\newtcolorbox{takeaways}[1][]{%
  enhanced,
  breakable,
  colback=white,
  colframe=bgcolor,
  arc=2pt,
  boxrule=0.6pt,
  left=0.9em,
  right=0.9em,
  top=0.7em,
  bottom=0.7em,
  title={\textbf{Takeaways}},
  colbacktitle=bgcolor,
  coltitle=fgcolor,
  toptitle=0.22cm,
  bottomtitle=0.12cm,
  borderline west={1.6pt}{0pt}{bgcolor},
  before skip=0.6em,
  after skip=0.6em,
  #1%
}

\def\alphabet{a,b,c,d,e,f,g,h,i,j,k,l,m,n,o,p,q,r,s,t,u,v,w,x,y,z}
\def\Alphabet{A,B,C,D,E,F,G,H,I,J,K,L,M,M,O,P,Q,R,S,T,U,V,W,X,Y,Z}
\def\greekalphabet{alpha,beta,gamma,delta,epsilon,varepsilon,zeta,eta,theta,vartheta,iota,kappa,varkappa,lambda,mu,nu,xi,pi,varpi,rho,varrho,sigma,varsigma,tau,upsilon,phi,varphi,chi,psi,omega}
\def\GreekAlphabet{Gamma,Delta,Theta,Lambda,Xi,Pi,Sigma,Upsilon,Phi,Psi,Omega}

\makeatletter
\def\changegreek{\@for\next:=\greekalphabet
	\do{\expandafter\let\csname\next\expandafter\endcsname\csname\next up\endcsname}}
\def\changegreekbf{\@for\next:=\greekalphabet
	\do{\expandafter\def\csname\next\expandafter\endcsname\expandafter{%
			\expandafter\bm\expandafter{\csname\next up\endcsname}}}}
\makeatother

\foreach \x in \alphabet {
	\expandafter\xdef\csname v\x\endcsname{\noexpand\ensuremath{\noexpand\constantvector{\x}}}

	\expandafter\xdef\csname ev\x\endcsname{\noexpand\ensuremath{\noexpand\x}}

	\ifthenelse{\equal{\x}{m}}{}{
		\expandafter\xdef\csname r\x\endcsname{\noexpand\ensuremath{\noexpand\randomscalar{\x}}}
	}

	\expandafter\xdef\csname rv\x\endcsname{\noexpand\ensuremath{\noexpand\randomvector{\x}}}
}

\foreach \x in \greekalphabet {
	\expandafter\xdef\csname v\x\endcsname{\noexpand\ensuremath{\noexpand\constantvector{\csname \x\endcsname}}}

	\expandafter\xdef\csname ev\x\endcsname{\noexpand\ensuremath{\noexpand{\csname \x \endcsname}}}

	\expandafter\xdef\csname r\x\endcsname{\noexpand\ensuremath{\noexpand\randomscalargreek{\csname \x\endcsname}}}

	\expandafter\xdef\csname rv\x\endcsname{\noexpand\ensuremath{\noexpand\randomvectorgreek{\csname \x\endcsname}}}
}

\foreach \x in \Alphabet {
	\expandafter\xdef\csname m\x\endcsname{\noexpand\ensuremath{\noexpand\constantmatrix{\x}}}

	\expandafter\xdef\csname em\x\endcsname{\noexpand\ensuremath{\noexpand\x}}

	\expandafter\xdef\csname rm\x\endcsname{\noexpand\ensuremath{\noexpand\randommatrix{\x}}}

	\expandafter\xdef\csname t\x\endcsname{\noexpand\ensuremath{\noexpand\tensorstyle{\x}}}

	\expandafter\xdef\csname g\x\endcsname{\noexpand\ensuremath{\noexpand\graphstyle{\x}}}

	\ifthenelse{\equal{\x}{E}}{}{
		\expandafter\xdef\csname s\x\endcsname{\noexpand\ensuremath{\noexpand\setstyle{\x}}}
	}

}

\foreach \x in \GreekAlphabet {
	\expandafter\xdef\csname m\x\endcsname{\noexpand\ensuremath{\noexpand\constantmatrixgreek{\csname \x\endcsname}}}

	\expandafter\xdef\csname rm\x\endcsname{\noexpand\ensuremath{\noexpand\randommatrixgreek{\csname \x\endcsname}}}
}

\newcommand{\E}{\mathbb{E}}
\newcommand{\Ls}{\mathcal{L}}
\newcommand{\R}{\mathbb{R}}

\title{The Design Space of Tri-Modal Masked Diffusion Models}

\author{Louis Bethune}
\author{Victor Turrisi}
\author{Bruno Kacper Mlodozeniec$^{3\dagger}$}
\author{Pau Rodriguez Lopez}
\author{Lokesh Boominathan}
\author{Nikhil Bhendawade}
\author{Amitis Shidani}
\author{Joris Pelemans}
\author{Theo X. Olausson$^{4\dagger}$}
\author{Devon Hjelm}
\author{Paul Dixon}
\author{João Monteiro}
\author{Pierre Ablin}
\author{Vishnu Banna}
\author{Arno Blaas}
\author{Nick Henderson}
\author{Kari Noriy}
\author{Dan Busbridge}
\author{Josh Susskind}
\author{Marco Cuturi}
\author{Irina Belousova}
\author{Luca Zappella}
\author{Russ Webb}
\author{Jason Ramapuram$^2$}

\affiliation{Apple, $^2$Google Deepmind (work done at Apple), $^3$University of Cambridge, $^4$MIT, $^{\dagger}$Work done during Apple internship.}

\abstract{
Discrete diffusion models have emerged as strong alternatives to autoregressive language models, with recent work initializing and finetuning a base unimodal model for bi-modal generation. Diverging from previous approaches, we introduce the first tri-modal \gls{mdm} \emph{pretrained from scratch} on text, image-text, and audio-text data. We systematically analyze multimodal scaling laws, modality mixing ratios, noise schedules, batch-size effects and provide optimized inference sampling defaults. 
Our batch-size analysis yields a novel
\gls{sde} based reparameterization, eliminating the need for tuning the optimal batch size as reported 
in recent work.
This re-parameterization 
decouples the \textit{physical batch size}, often chosen based on compute (GPU saturation, FLOP-efficiency, wall-clock time) from the \textit{logical batch size}, chosen to balance the variance of gradients during stochastic optimization.
Finally, we pretrain a preliminary model showcasing the capabilities of a unified design, achieving strong results at 3B model scale (6.4T tokens), in both text generation, T2I tasks, and T2S tasks.
Our work represents the largest scale systematic open study of multimodal discrete diffusion models conducted to date, providing valuable insights into scaling behaviors across multiple modalities.
}

\metadata[Correspondence]{\sffamily
Louis Bethune: \url{louisbethune@apple.com};
Victor Turrisi: \url{v_turrisi@apple.com};
Jason Ramapuram: \url{jramapuram@google.com}. For a detailed breakdown of author contributions see~\Cref{sec:contributions}.
}

\begin{document}
\maketitle
\section{Introduction}
\label{sec:introduction}

A recurring theme in sequence modeling is that, whenever the full context is available, bidirectional information tends to perform better. Early work on bidirectional RNNs \citep{DBLP:journals/tsp/SchusterP97} and LSTMs \citep{DBLP:journals/nn/GravesS05} demonstrated clear gains over purely forward recurrent models when both past and future states were accessible during training. This makes the dominance of causal transformers \citep{DBLP:conf/nips/VaswaniSPUJGKP17} in modern language modeling slightly surprising: the strongest transformer-based language models \citep{DBLP:journals/corr/abs-2412-19437,DBLP:journals/corr/abs-2312-11805,singh2025openai,claude3} are trained with a strictly left-to-right factorization (auto-regressively). The causal constraint is \emph{undeniably convenient} (simple likelihood factorization, efficient per-token learning signal and fast streaming decoding via KV cache), but it is not evidently the best fit for conditional generation problems where the observed evidence may be scattered across positions and modalities. Discrete diffusion revisits the bidirectional viewpoint by replacing a fixed generation order with iterative refinement: rather than predicting the next token, the model repeatedly denoises a partially corrupted sequence. Recent progress in diffusion based language modeling \citep{DBLP:journals/corr/abs-2502-09992} has narrowed much of the quality gap to strong causal baselines, including widely used models such as LLaMA~3~\citep{DBLP:journals/corr/abs-2407-21783}, strengthening the case that order agnostic, globally conditioned generation can be competitive at scale. Despite narrowing the performance gap at equivalent pretraining FLOP budgets, naive implementations still exhibit substantial latency compared to autoregressive baselines and further work is required to improve sampling efficiency in \gls{mdm} \citep{jazbec2025learning,DBLP:journals/corr/abs-2505-22618}.  

This refinement perspective is particularly appealing in multimodal settings. \gls{mdm}s train on a simple corruption process (masking) and learn to reconstruct missing tokens. With multimodal tokenization, text, image, and audio tokens can be concatenated, 
partially masked, and jointly denoised. This naturally supports infilling and arbitrary conditioning without re-deriving a new factorization for every task. 
Although image and audio tokens could be independently modeled in continuous domains as in \citet{DBLP:journals/corr/abs-2509-16197}, we instead adopt discrete modeling to streamline optimization and substantially reduce complexity by employing a unified embedding space and loss function.

While much of the current multimodal \gls{mdm} literature emphasizes adapting existing pretrained models, either by performing supervised finetuning on discrete diffusion bases like LLaDA \citep{DBLP:journals/corr/abs-2505-16933, DBLP:journals/corr/abs-2505-15809} or by distilling and repurposing autoregressive backbones such as Qwen 2.5-Coder or other large AR models \citep{DBLP:journals/corr/abs-2409-12186,DBLP:journals/corr/abs-2506-20639, DBLP:journals/corr/abs-2510-01329, DBLP:conf/icml/ZhangZ0TOSJ25,DBLP:journals/corr/abs-2512-15745},
our work targets the pretraining regime, where the dominant compute is spent and where the latent spaces are shaped.
\begin{figure*}[t]
    \centering
    \begin{subfigure}[t]{0.24\textwidth}
        \centering
        \includegraphics[width=\linewidth]{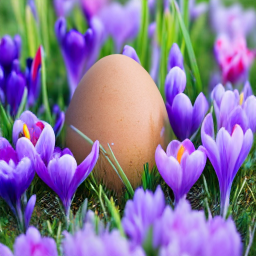}

        \caption{}
    \end{subfigure}
    \hfill
    \begin{subfigure}[t]{0.24\textwidth}
        \centering
        \includegraphics[width=\linewidth]{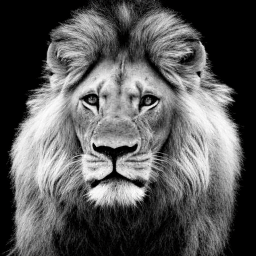}
        \caption{}
    \end{subfigure}
    \hfill
    \begin{subfigure}[t]{0.24\textwidth}
        \centering
        \includegraphics[width=\linewidth]{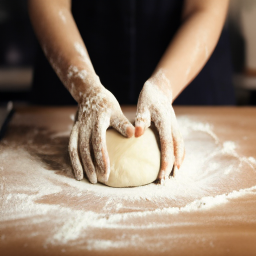}
        \caption{}
    \end{subfigure}
    \hfill
    \begin{subfigure}[t]{0.24\textwidth}
        \centering
        \includegraphics[width=\linewidth]
        {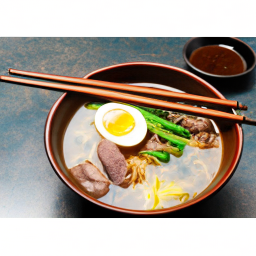}        
        \caption{}
    \end{subfigure}
    \caption{\textbf{High-fidelity generation.} The pretrain-only 3B \gls{mdm} demonstrates strong prompt adherence alongside high-quality visual rendering of \textit{texture}, \textit{lighting}, and \textit{composition}. Samples show: 
             \textbf{(a)} natural daylight and depth of field ("egg in a field of crocuses"); 
             \textbf{(b)} fine-grained fur texture in B\&W ("lion's face"); 
             \textbf{(c)} soft, warm lighting with vintage color tones ("preparing bread dough"); and 
             \textbf{(d)} complex multi-object arrangement ("noodle soup with toppings"). 
             Extended generations in Appendix \Cref{app:extended_generations}.
    }
    \label{fig:image_generations}
\end{figure*}
Moving from unimodal to multimodal \gls{mdm} introduces a large and underexplored design space. Choices that may appear secondary in isolation  %
can dominate stability and compute efficiency. Because exhaustive large-scale sweeps are often infeasible, progress depends on reliable transfer rules, i.e. rules on how to transfer hyperparameters from small models to larger models. %

In this work, we take a step toward sound pretraining recipes for native multimodal \gls{mdm}s. We extend \gls{mdm} to a tri-modal setting (text, image, audio) via a unified discrete token space with modality-specific boundary and mask tokens. %
The \emph{single} resulting model supports multiple conditional generation queries, including text-to-image generation, image captioning, \gls{tts} and \gls{asr}.

Our main contribution is an empirical study of the pretraining and inference choices that govern scaling behavior and efficiency in this regime, together with controlled inference-time ablations that reveal modality-specific sensitivities:%
\begin{enumerate}
    \item \textbf{Unified Multimodal MDM.} Introduce the first tri-modal \gls{mdm} capable of generating text, image, and audio from each other \emph{in any direction} via a single transformer backbone and unified vocabulary, eliminating the need for modality-specific heads, adapters or unimodal conditioning \citep{DBLP:conf/nips/LiuLWL23a}.
    \item \textbf{Elimination of Optimal Batch Size ($B_{\text{opt}}$) and per-module transfer.} We leverage SDE-based reparameterization to render training loss invariant to batch size up to a critical threshold ($B_\text{crit}$), \emph{eliminating the need to search for an optimal batch size}, $B_{\text{opt}}$ \citep{DBLP:journals/corr/abs-2505-13738}.
    We also validate the effectiveness of per-module (e.g., MLP, attention weights) hyperparameter scaling using CompleteP + SDE scaling~\citep{mlodozeniec2025completed} within the multimodal \gls{mdm} regime (\autoref{app:per_module_hyperparam}).
    \item \textbf{Multimodal Scaling Laws.} We derive empirical scaling laws for validation loss as a function of model size ($N$) and token budget ($D$), providing prescriptive guidance for compute optimal tri-modal MDMs. 
    We find the seminal formula $L(N, D)=E+(AN^{-a/b}+BD^{-1})^b$ from \citet{DBLP:journals/corr/abs-2001-08361} to be a better fit than the additive form of~\citet{DBLP:journals/corr/abs-2203-15556}. 
    In particular, we find these models to be asymptotically more data efficient than their auto-regressive counterpart, with the compute optimal
    frontier of $D^{\star}(N)\approx 7754\cdot N^{0.84}$.
    \item \textbf{Modality Dependent Design Space.} We characterize the distinct inference tradeoffs for each modality, identifying that optimal noise schedules and sampling parameters (guidance, temperature) differ significantly between text, image, and audio generation.
\end{enumerate}

\section{Background and Related Work}
\label{sec:background}

\subsection{Masked Diffusion Models}
\label{sec:discretediff_unimodal}

Although diffusion models first gained prominence through their success in continuous settings such as image generation~\citep{DBLP:conf/iclr/SongME21}, the original formulation by \citet{DBLP:conf/icml/Sohl-DicksteinW15} provided a unified framework encompassing both continuous and discrete domains.
One form of diffusion on discrete data, \gls{mdm}s were first proposed by \citet{DBLP:conf/nips/AustinJHTB21}, and generalized in earlier discrete diffusion work by \citet{DBLP:conf/nips/HoogeboomNJFW21}. 
In their formulation, the diffusion forward steps, $q(x_t|x_{t-1})$, progressively noise the data $x_0$ with mask tokens [MASK], turning the data distribution $q_0 := p_{data}$ into a stationary distribution $q_T := q(x_T)$ 
in which every token is masked.
A masked diffusion model $p_{\theta}(x_{t-1}|x_t)$ with parameters $\theta$ then learns the reverse process such that starting from the stationary masked distribution $q_T$, the reverse process $ \prod_t p_{\theta}(x_{t-1}|x_t)$ reconstructs the original text from such noised sequences. %
This approach for masked diffusion with fixed timesteps $t \in 0,1,\ldots,T$ was extended to a continuous-time framework by \citet{DBLP:conf/nips/CampbellBBRDD22}, who formulated the forward noising process and corresponding reverse process as Continuous Time Markov Chains (CTMCs), and by \citet{DBLP:journals/corr/abs-2211-15089}, who propose embedding discrete inputs into continuous space before learning the diffusion model.  

Below, we provide an overview of applications of MDMs to the three modalities we tackle in this work.

\paragraph{Text.} \citet{DBLP:conf/nips/AustinJHTB21} first applied MDMs to relatively small-scale text datasets like LM1B \citep{DBLP:conf/interspeech/ChelbaMSGBKR14}. Then, recent adaptations of the continuous-time framework by \citet{DBLP:conf/nips/CampbellBBRDD22} have enabled training of larger language diffusion models. LLaDA \citep{DBLP:journals/corr/abs-2502-09992}, for instance, trained an 8B-parameter \gls{mdm} on a 2.3T token corpus, obtaining strong performance on benchmarks such as MMLU \citep{DBLP:conf/iclr/HendrycksBBZMSS21} and GSM8K \citep{DBLP:journals/corr/abs-2110-14168}. In contrast, Dream \citep{DBLP:journals/corr/abs-2508-15487} finetuned a pretrained autoregressive Qwen2.5-7B model using a 580B token corpus \citep{DBLP:journals/corr/abs-2412-15115}, without accounting for the initial AR model's pretraining budget. Both methods employ the same training objective, representing an upper bound on the negative log-likelihood, or \gls{elbo}, of the continuous-time masked diffusion process:
\begin{equation}\label{eq:elbo-mdm}
    - \mathbb{E}_{\substack{x_0 \sim p_{data}, ~ t \sim U(0,1) \\ x_t \sim q(x_t|x_0) }}\left[w(t) \sum_{\ell=1}^L{\mathbf{1}_{x^\ell_t = \text{MASK}}\, p_{\theta}(x_0^\ell|x_t)}\right] .
\end{equation}
Here $L$ is the sequence length and $x^\ell$ denotes the $\ell-$th token of $x$. The indicator function $\mathbf{1}_{x^\ell_t = \text{MASK}}$ makes sure the loss is computed only over masked tokens. The weights $w(t)$ depend on the form of the forward noise $q(x_t|x_{s})$; for LLaDA, it is $w(t) = 1/t$, while Dream uses a more complex schedule. 
Follow up works such as \citet{DBLP:journals/corr/abs-2505-19223} and \citet{DBLP:journals/corr/abs-2509-24389} improve performance and efficiency by introducing variance reduction and mixture-of-experts (MoE) methods to large language diffusion. We provide a principled exposition of weighting in \autoref{app:weighting}.

\paragraph{Image.}  
\citet{DBLP:conf/nips/AustinJHTB21} and \citet{DBLP:conf/nips/ShiHWDT24} apply masked diffusion directly to pixel values by modeling them as categorical variables. However, their experiments are restricted to low resolution image datasets, such as CIFAR10 and downsampled ImageNet 64x64 \citep{DBLP:conf/cvpr/DengDSLL009, DBLP:conf/icml/OordKK16}.  
MaskGIT and VQ-Diffusion  ~\citep{DBLP:conf/cvpr/ChangZJLF22, DBLP:conf/cvpr/GuCBWZCYG22} instead use pretrained image tokenizers such as VQ-GAN~\citep{DBLP:conf/cvpr/EsserRO21, sber_movqgan, zheng2022movqmodulatingquantizedvectors} or VQ-VAE~\citep{DBLP:conf/nips/OordVK17} to downsample the full set of image pixels into a patch-wise grid of discrete tokens, enabling stable high-resolution discrete image diffusion training.

\paragraph{Audio.} Literature around speech and audio generation is generally scarcer. \citet{DBLP:journals/taslp/YangYWWWZY23} apply discrete diffusion to audio generation. 
\citet{DBLP:journals/corr/abs-2305-09636} combine the SoundStream audio tokenizer~\citep{DBLP:journals/taslp/ZeghidourLOST22} with a masking-unmasking approach similar to MaskGIT~\citep{DBLP:conf/cvpr/ChangZJLF22} for audio generation. %

\subsection{Multimodal Masked Diffusion Models}
\label{sec:discretediff_multimodal}
Some elements of multimodality were introduced by works such as \citet{DBLP:conf/cvpr/GuCBWZCYG22} (text-to-image generation) and \citet{DBLP:journals/corr/abs-2505-16933} (visual question answering). However, these models are still restricted to generating only one modality.
In contrast, by applying a unified probabilistic formulation that represents heterogeneous data as a single stream of concatenated discrete tokens, MMaDA \citep{DBLP:journals/corr/abs-2505-15809} unifies language modeling, image understanding, and image generation as a multimodal \gls{mdm}. MMaDA is initialized from LLaDA's weights and subsequently trained with the same objective as LLaDA and Dream in \Cref{eq:elbo-mdm} but on the joint
token stream of image and text tokens. 
\citet{DBLP:journals/corr/abs-2503-20853} and \citet{DBLP:conf/iclr/0001ZYCZWTS23} train a unified discrete diffusion model for both image and text at smaller scale, with the latter using a mix of masked and uniform-state diffusion.
We move beyond existing bi-modal (text–image) MDMs by introducing audio as a novel third modality, addressing new multimodal challenges. Unlike~\citet{DBLP:journals/corr/abs-2505-15809}, we \emph{pretrain from scratch}, properly account for the total token budget, and jointly optimize representations across all modalities. We encourage the community to jointly report model and data size, $(N, D_{\text{total}})$, to make fair comparisons between hypothesis classes.

\begin{figure}[t]
    \centering
    \begin{minipage}[t]{0.48\linewidth}
        \centering
        \includegraphics[width=0.98\linewidth]{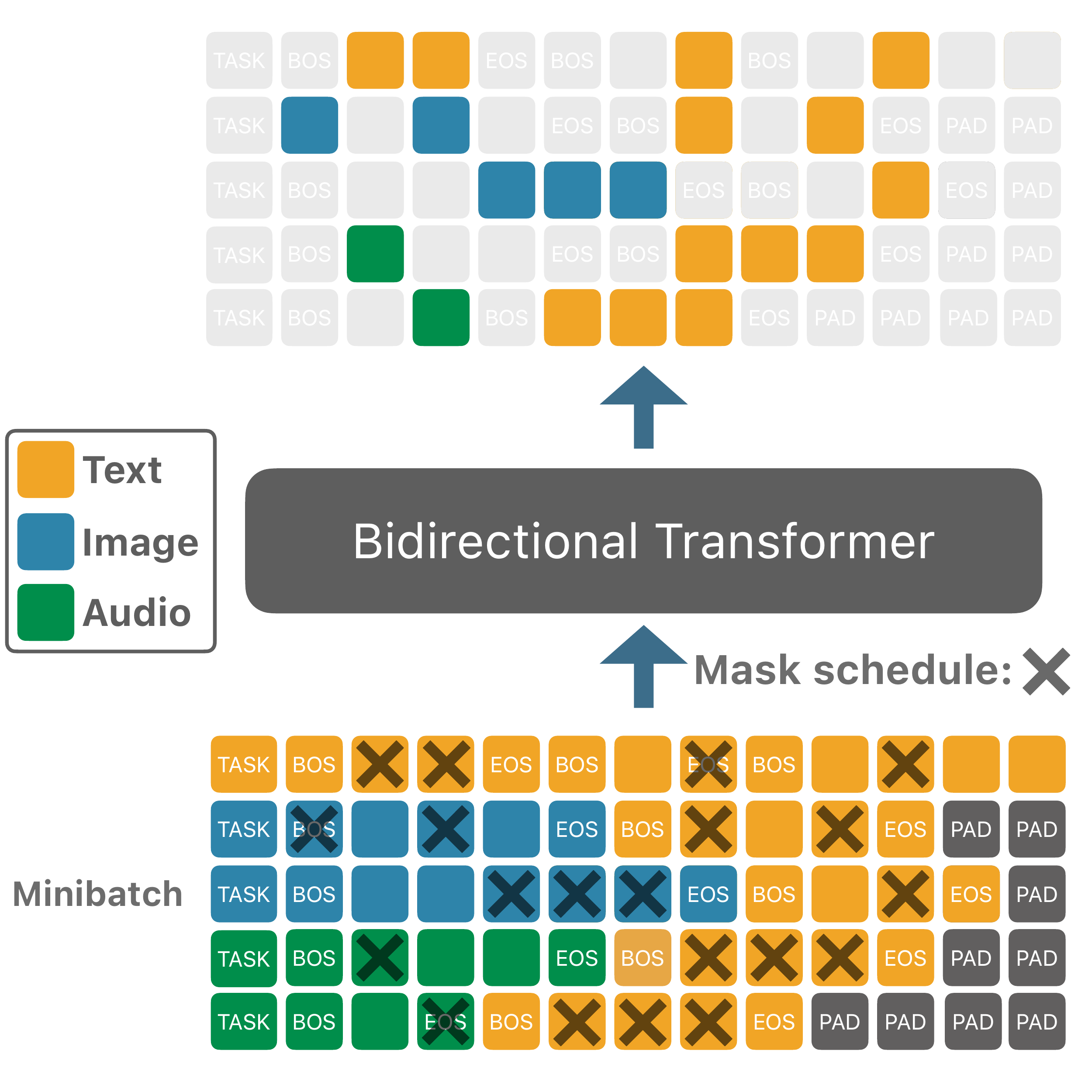}
        \caption{\textbf{Tri-Modal masked diffusion model architecture}. 
        Pure text is packed. Image-caption and audio-transcription pairs are padded to maximum length. Padding is ignored by attention and loss computation.}
        \label{fig:lisa-method}
    \end{minipage}\hfill
    \begin{minipage}[t]{0.48\linewidth}
        \centering
        \includegraphics[width=1.12\linewidth]{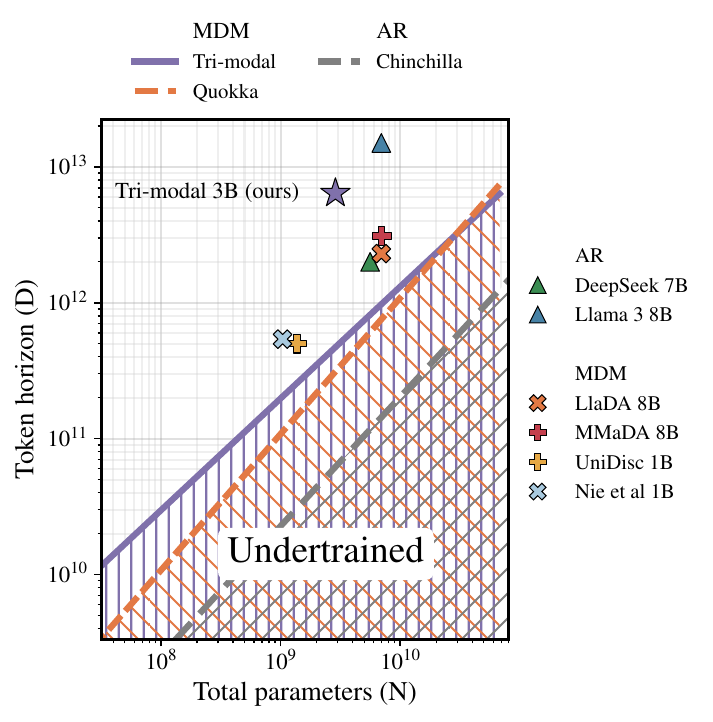}
        \caption{\textbf{Token-optimal curve $D^{\star}(N)$ for different model families}. In  tri-modal MDM, token count growth sub-linearly with model size, suggesting diminishing returns of additional data. We use identical methodology to report all curves.}
        \label{fig:pareto-front-flops}
    \end{minipage}
\end{figure}

\section{Method}
\label{sec:method}

We consider three modalities $m \in \gM := \{\text{text}, \text{audio}, \text{image}\}$. Each training sample belongs to one of three categories: text-only, image-text pairs or audio-text pairs, where each modality is represented as a sequence of discrete tokens drawn from a modality-specific vocabulary,\looseness=-1
\begin{align*}
    x_m = (x_m^1, \ldots, x_m^{L_m}), \qquad x_m^i \in \gV_m,
\end{align*}
where $|\gV_m| = V_m$. To enable unified modeling, we construct a shared vocabulary $\gV = \gV_{\text{text}} \sqcup \gV_{\text{audio}} \sqcup \gV_{\text{image}}$, where \(\sqcup\) denotes disjoint union. We further introduce modality-specific special tokens
\begin{align*}
\gV_{\text{spec}} = \{\text{BOS}_m, \text{EOS}_m, \text{MASK}_m \;:\; m \in \gM\},
\end{align*}
as well as a padding token \(\text{PAD}_{\text{text}}\) that is added after the text prompt of multimodal sequences that are shorter than our target sequence length. Lastly, we introduce three special task tokens
\begin{align*}
\gV_{\text{task}} = \{\text{TASK}_\text{text}, \text{TASK}_\text{image-text}, \text{TASK}_\text{audio-text}\},
\end{align*}
that allows us to signal to the model which task it needs to perform. This is especially useful if one wishes to add new tasks to the model through mid-training or supervised finetuning. The resulting unified vocabulary has size $\lvert V \rvert =  \sum_{m \in \gM} \lvert V_m \rvert + \lvert V_{\text{spec}} \rvert + \lvert V_{\text{task}} \rvert + 1$. 

To avoid any confusion regarding modalities and diffusion time, we switch our notation from the generic signal $x$ to the full training sequence $s$. Throughout the paper, superscripts denote position indices (i.e., token positions), while subscripts denote diffusion time or modality indices.

\paragraph{Training.} Training sequences are constructed by wrapping modality tokens with their boundary tokens. For example, an audio-text sample is represented as:
\vspace{-0.1cm}
\begin{align*}
    s = [&\text{TASK}_\text{audio-text}, \text{BOS}_{\text{audio}}, x_{\text{audio}}^1, \ldots, x_{\text{audio}}^{L_{\text{audio}}}, \text{EOS}_{\text{audio}},
     \text{BOS}_{\text{text}}, x_{\text{text}}^1, \ldots, x_{\text{text}}^{L_{\text{text}}}, \text{EOS}_{\text{text}}],
\end{align*}
with an image-text sample following the same format. Since we train with maximum sequence length $L^\star$, text-only sequences are packed and truncated so that they always have exactly $L^\star$ positions, i.e., no padding is necessary. On the other hand, mixed-modality sequences whose total length is shorter than $L^\star$ are right-padded after \(\text{EOS}_{\text{text}}\) using the token \(\text{PAD}_{\text{text}}\) to match $L^\star$. See \Cref{fig:lisa-method} for a demonstration.

Given a sequence $s = (s^1, \ldots, s^{L^\star}) \in \gV^{L^\star}$, we define a continuous-time forward masking process indexed by $t \in [0,1]$. Each position is independently corrupted according to a Bernoulli masking mechanism with probability $\beta_t$, where $\beta_{\cdot}$ denotes a monotonic function with $\beta_0 = 0$ and $\beta_1 = 1$. Let $m(i) \in \mathcal{M}$ denote the modality associated with position $i$. The corrupted token $s_t^i$ is sampled as
\vspace{-0.1cm}
\begin{align*}
    s_t^i | s_{t-1}^{i} \sim
\begin{cases}
\text{MASK}_{m(i)} & \text{with probability } \beta_t,\\
s_{t-1}^i               & \text{with probability } 1 - \beta_t.
\end{cases}
\end{align*}
This defines a corrupted sequence $s_t = (s_t^1,\ldots,s_t^{L^\star})$ with $s_0 = s$. Masking is applied independently across positions and modalities, with each modality using its own dedicated mask token $\text{MASK}_m$. Note that the task tokens are never masked. The parameter $t$ controls the overall corruption level, smoothly interpolating between the original sequence at $t = 0$ and a fully masked sequence at $t = 1$. For a position \(i\) associated with modality \(m(i)\), the forward noising kernel is given by
\begin{align*}
    q(s_t^i \mid s^i) = \bar{\beta}_t \cdot \delta_{\text{MASK}_{m(i)}}(\rvs_t^i) + (1-\bar{\beta}_t) \cdot \delta_{\rvs^i}(\rvs_t^i)\,,
\end{align*}
where \(\delta_v(\cdot)\) denotes the Dirac measure centered at \(v\), and $\bar{\beta}_t = \prod_{ t' \leq t} \beta_{t'}$. Since masking is applied independently across positions, the forward process factorizes as
\vspace{-0.2cm}
\begin{align*}
    q(s_t \mid s)
    = \prod_{i=1}^{L^\star} q(s_t^i \mid s^i).
\end{align*}
Equivalently, this induces a Markov kernel between successive noise levels
\vspace{-0.2cm}
\begin{align*}
    q(s_t \mid s_{t-1})
            = \prod_{i=1}^{L^\star}
    \Bigl[
        \alpha_t \, \delta_{s_{t-1}^i}(s_t^i)
        + (1-\alpha_t)\, \delta_{\text{MASK}_{m(i)}}(s_t^i)
    \Bigr],
\end{align*}
where $\alpha_t = 1 - \beta_t$ is chosen such that the marginal distribution of $s_t$ matches $q(s_t \mid s)$ (see \Cref{app:weighting} for more details). The monotonic nature of the masking process is clear: once a token is masked, it stays masked.

\paragraph{Denoising model.} We parameterize the reverse process using a denoising model $f_\theta: \gV^{L^\star} \to \R^{L^\star \times V}$ which predicts logits over the unified vocabulary at each position. Given a corrupted sequence $s_t$, the model outputs $h = f_\theta(s_t)$, where $h^i_v$ denotes the logit assigned to token $v \in \gV$ at position $i$, and the ground-truth token $s^i$, we define the per-token loss as
\vspace{-0.2cm}
\begin{align*}
    \ell_i(\theta, s)
    = -\log
    \frac{\exp\!\big(h^i_{s^i}\big)}
    {\sum_{v \in \gV} \exp\!\big(h^i_v\big)} := -\log p_\theta(s^i \mid s_t).
\end{align*}
For memory efficiency and to enforce modality constraints, we implement this loss using cut-cross-entropy (CCE) \citep{DBLP:journals/corr/abs-2411-09009}, which avoids materializing the full probability distribution.

Let $\gI_t := \{\, i \mid s_t^i = \text{MASK}_{m(i)} \,\}$ denote the set of masked, non-padding positions at time $t$. The training objective is
\begin{align*}
\Ls(\theta)
=
\E_{\substack{s \sim \mathcal{D},\\ t \sim \mathrm{U}(\epsilon,1)}}
\left[
\frac{w(t)}{|\gI_t|}
\sum_{i \in \gI_t} \ell_i(\theta, s)
\right]
\;+\;
\lambda\,\Ls_z,
\end{align*}
where \(\epsilon,\lambda > 0\) are small constants for numerical stability, and \(\Ls_z\) denotes the z-loss regularizer~\citep{DBLP:journals/corr/BrebissonV16a}. We follow prior work~\citep{DBLP:journals/corr/abs-2502-09992, DBLP:journals/corr/abs-2505-15809} on masked diffusion models, and choose $w(t) = 1/t$ which yields an unbiased estimator of the ELBO under Bernoulli masking. See \Cref{app:weighting} for the effect of this weighting scheme and a more general formulation.

\paragraph{Inference.} At generation time, we iteratively unmask tokens according to a predefined linear schedule. Multimodal generation is conditioned on a prompt, e.g., text, and a target modality $m \in \gM$. We start the process from a fully masked sequence of the form
\vspace{-0.2cm}
\begin{align*}
s_K = [&\text{TASK}_\text{task}, \text{BOS}_m, \text{MASK}_m, \ldots, \text{MASK}_m, \text{EOS}_m, \text{BOS}_{\text{text}}, x_{\text{text}}^1, \ldots, x_{\text{text}}^{L_{\text{text}}}, \text{EOS}_{\text{text}}].
\end{align*}
For text-only generation, we instead construct the fully masked sequence of the form
\vspace{-0.2cm}
\begin{align*}
s_K = [&\text{TASK}_\text{text}, \text{BOS}_{\text{text}}, x_{\text{text}}^1, \ldots, x_{\text{text}}^{L_{\text{text}}}, \text{MASK}_{\text{text}}, ..., \text{MASK}_{\text{text}}, \text{EOS}_{\text{text}}].
\end{align*}
At each reverse diffusion step $k \in [K]$, where $K$ denotes the number of generation steps, the denoising model produces $h_k = f_\theta(s_{k-1})$, where $h_{k}^i \in \R^{\gV}$ denotes the logits at position $i$. For each masked position $i$, a candidate token is sampled from the modality-constrained predictive distribution
\vspace{-0.2cm}
\begin{align*}
    s_k^i \sim p_\theta(\,\cdot \mid s_{k-1})
\;\;\propto\;\;
\exp\!\big(h_{k,v}^i\big),
\qquad v \in \gV_{m(i)}.
\end{align*}
Based on the unmasking schedule, a subset of masked positions are revealed, updating the sequence $s_k$. The process is repeated until no masked positions remain, producing the final generated sample.
\subsection{Architecture}
\label{sec:architecture}
For all experiments presented in this paper, 
we rely on a standard bi-directional transformer architecture with pre-normalization RMSNorm~\citep{DBLP:conf/nips/ZhangS19a}, SwiGLU MLPs~\citep{DBLP:journals/corr/abs-2002-05202}, rotary positional embeddings (RoPE)~\citep{DBLP:journals/corr/abs-2104-09864} and QK-norm~\citep{DBLP:conf/icml/0001DMPHGSCGAJB23,DBLP:conf/iclr/WortsmanLXEAACG24,DBLP:journals/corr/abs-2405-09818, DBLP:journals/corr/abs-2503-19786}. 
Our Tri-modal 3B model is pretrained from scratch for 1M steps with batch size of $3072$ and sequence length of $3256$. 
We tokenize modalities with specialized encoders: SBER-MoVQGAN~\citep{sber_movqgan} for images, Higgs Audio v2~\citep{higgsaudio2025} for audio, and Tiktoken~\citep{openai_tiktoken} for text. To manage the large vocabulary efficiently, we employ cut-cross-entropy~\citep{DBLP:conf/iclr/WijmansHHKK25} and apply a z-loss regularizer~\citep{DBLP:journals/corr/BrebissonV16a} to stabilize logits amplitudes. See \autoref{tab:training_details} for full hyperparameter details.

\section{Hyperparameter Transfer}
\label{sec:hyperparameter_transfer}
Selecting optimal hyperparameters is of paramount importance to the final performance of the model and conducting a grid search at large scale is not feasible. In this work, we rely on \textit{hyperparameter transfer rules} to transfer the optimal set found at small scale to larger scale. Several rules have been proposed in the literature: $\mu$P~\citep{yang2022tensor} proposed a scaling for width, later extending to depth with depth-$\mu$P~\citep{yang2023tensor}, u-$\mu$P~\citep{blakeu} and CompleteP~\citep{dey2025don}. Here, we rely on the work of~\citet{mlodozeniec2025completed}, an extension of CompleteP~\citep{dey2025don}.
Additionally, \citet{mlodozeniec2025completed} recently demonstrated performance gains from per-module hyperparameter optimization, adjusting multipliers for AdamW parameters (learning rate, weight decay, momenta $\beta_1$, $\beta_2$, and $\epsilon$) across distinct modules like MLP weights, attention projections, embeddings, and normalization layers. Our work provides the first empirical validation of this refined tuning in the context of multimodal MDMs, with preliminary results presented in \autoref{app:per_module_hyperparam}.

\subsection{Eliminating $B_{\text{opt}}$ with SDE Parametrization}
\label{sec:b_opt}
\begin{figure}[th]
    \centering
    \begin{minipage}[t]{0.49\linewidth}
        \centering
        \includegraphics[width=\linewidth]{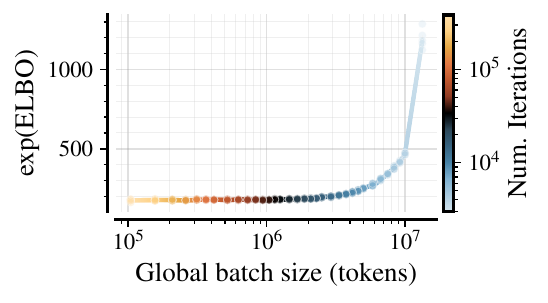}
        \caption{\textbf{Below the critical batch size $B_{\text{crit}}$ the SDE parametrization guarantees constant loss}. In that regime, larger batch sizes allow fewer iterations. Above it, SDE discretization breaks and training ceases to be FLOP-efficient.}
        \label{fig:boptflat}
    \end{minipage}\hfill
    \begin{minipage}[t]{0.49\linewidth}
        \centering
        \includegraphics[width=\linewidth]{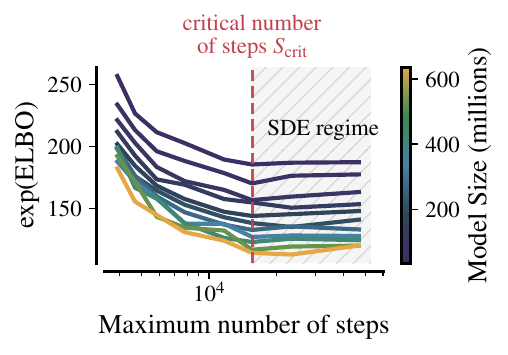}
        \caption{\textbf{Critical iteration count $S_{\text{crit}}$ is constant w.r.t.\ model size} under the SDE regime. This is compatible with the findings of~\citet{DBLP:journals/corr/abs-2505-13738}, but their study was done \textit{outside} the SDE regime.}
        \label{fig:ScritfromN}
    \end{minipage}
\end{figure}

Stochastic Optimization with AdamW operates like the discretization of a Stochastic Differential Equation (SDE)~\citep{DBLP:conf/nips/MalladiLPA22,mlodozeniec2025completed} whose timescale, noise, and drift, can be computed from AdamW's parameters. According to these studies, AdamW's hyperparameters are redundant with batch size: for example, it is possible to reduce the noise in the gradient estimation, either by increasing the batch size or by decreasing the momentum weight. Similarly, lower noise allows for larger step sizes. Since batch size is typically constrained by the compute budget and the memory available on the chip, it is desirable to make it a free hyperparameter whose value does not interfere with the performance of the model. We thus re-parametrize the hyperparameters in batch-size with~\citet{mlodozeniec2025completed} to train the network with any batch size, guaranteeing similar performance across all compute budget, \textit{as long as the batch size is not larger than $B_{\text{crit}}$}. Results are illustrated in~\autoref{fig:boptflat}. SDE parametrization guarantees homogeneous behavior across batch sizes, including the smallest ones. This contrasts with the typical U-curve associated with non-SDE parametrization~\citep{DBLP:journals/corr/abs-2505-13738}, where there is a balance between total drift (when batch size is too big) and excessive noise (when batch size is too small).  

\subsection{Isonoise and Isohorizon Scaling Rules}
\label{sec:isonoisedef}
The SDE is the continuous limit of the gradient flow elicited by AdamW. Intuitively, the SDE horizon corresponds to the trajectory length in parameter space (extending from origin), while SDE drift controls the scale of stochastic fluctuations induced by gradient noise. We propose a new way to balance these two contributions, controlled by a parameter $\gamma\in[0,1]$. When increasing the number of tokens $D$, two quantities can be conserved: \textbf{(a)} we can conserve the SDE drift (isonoise curves) with $\gamma=0$, or \textbf{(b)} we can conserve the SDE horizon (isohorizon curves) with $\gamma=1$. 
We smoothly interpolate between these extremes for intermediate values ($0<\gamma<1$) by defining the SDE-scaling factor $\kappa$ as:
\begin{equation}
\kappa=\left(\frac{D^{\text{base}}}{D}\right)^{\gamma}\left(\frac{B}{B^{\text{base}}}\right),
    \label{eq:isonoisedef}
\end{equation}
where $D^{\text{base}}$ is the base model size, $D$ is the target model size, $B^{\text{base}}$ is the base batch size and $B$ the target batch size. Then, AdamW's hyperparameters are rescaled using $\kappa$ as:
\begin{equation}
    \text{lr}= \text{lr}^{\text{base}}\sqrt{\kappa},\qquad
    \beta_1 = (\beta_1^{\text{base}})^{\kappa},\qquad
    \beta_2 = (\beta_2^{\text{base}})^{\kappa},\qquad
    \epsilon = \frac{\epsilon^{\text{base}} }{\sqrt{\kappa}}.
    \label{eq:adamwnewformulation}
\end{equation}
We conduct an initial hyperparameter search of $(\text{lr}^{\text{base}},\beta_1^{\text{base}},\beta_2^{\text{base}},\epsilon^{\text{base}})$ with $\approx 3$k runs with a N=320M model (including 240M embedding parameters), $D^{\text{base}}=13$B tokens, and global batch size of $B^{\text{base}}=256$ sequences. %

\section{Scaling Behavior of \gls{mdm} under the SDE Transfer Rule}
\label{sec:scaling}

This section is devoted to the scaling properties of tri-modal \gls{mdm}, under the CompleteP + SDE reparametrization regime. 
All experiments presented use a cosine learning rate schedule with 1k steps of linear warmup, constrained so that warmup never exceeds 25\% of the total iteration count.
Following~\citet{DBLP:conf/icml/BusbridgeSWRLW25}, we set width proportional to depth, fixing $\rho = d_{\text{emb}} / n_{\text{layers}} = 128$ while scaling up models, ensuring consistent hyperparameter transfer and more stable scaling behavior.
\subsection{Scaling Rules for Critical Batch Size}
\label{sec:critical_batch_size}

\begin{figure}[th]
    \centering
    \begin{minipage}[t]{0.48\linewidth}
        \centering
        \includegraphics[width=\linewidth]{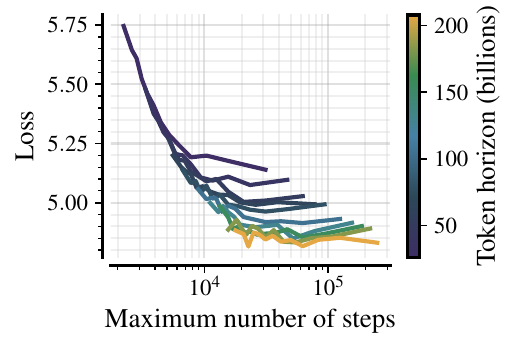}
        \caption{\textbf{Critical iteration count $S_{\text{crit}}$ increases with token horizon $D$}. The increase is sub-linear, meaning that the critical batch size $B_{\text{crit}}$ also increases with horizon $D$.}
        \label{fig:LfromScrit}
    \end{minipage}\hfill
    \begin{minipage}[t]{0.45\linewidth}
        \centering
        \includegraphics[width=\linewidth]{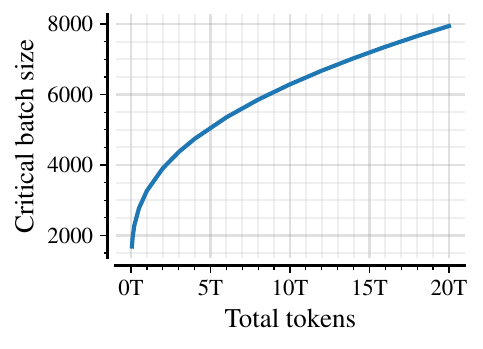}
        \caption{\textbf{Critical batch-size $B_{\text{crit}}$ as a function of the token horizon}. There is an intrinsic tension between wall-clock time and FLOP-efficiency.}
        \label{fig:BcritfromD}
    \end{minipage}
\end{figure}

\paragraph{Critical batch size without SDE.} When not using SDE parametrization, the batch size impacts both the variance of the stochastic gradient estimation (the SDE drift), and the iteration count $S$ (the SDE horizon). Practically, it means that beyond $B_{\text{crit}}$ the marginal utility of each additional token in the batch decreases.  
Previous work in AR~\citep{DBLP:journals/corr/abs-2001-08361,DBLP:journals/corr/abs-2412-19437} modeling fit a power law to enable predicting critical batch size in tokens $D$ or compute budget $C$. More recent work \citep{DBLP:journals/corr/abs-2505-13738} suggests there exists a $B_\text{opt}$: the batch size that minimizes the loss at a given token horizon $D$. Under SDE parametrization, that notion disappears, as shown in~\autoref{fig:boptflat}: all batch sizes under $S_{\text{crit}}$ yield identical results at fixed token budget $D$. See Appendix \Cref{sec:runtime_batch_size} for more details.

\paragraph{Critical batch size for SDE.} $S_{\text{crit}}$ is estimated empirically as the minimum number of optimization steps required to maintain FLOP-efficient training. When the number of integration steps $S_{\text{crit}}$ is too low, the SDE approximation breaks and the performance \textit{at constant horizon} $D$ plummets. This can be expressed in term of the critical batch size $B_{\text{crit}}=D/(L{S_{\text{crit}}})$ with $L$ being the sequence length. Above $S_{\text{crit}}$ the asymptotic loss depends mainly on the model size $N$ and token budget $D$, irrespective of iteration count. We illustrate this phenomenon in~\autoref{fig:boptflat} with a model of size 320M trained on 13B tokens. This means that below $S_{\text{crit}}$, all runs are \textit{FLOP-efficient}: they minimize the loss for the token budget. Above that, runs are \textit{wall-clock inefficient}: they trade faster training for wasteful usage of tokens.

\begin{takeaways}
Under SDE parameterization, the global batch size can be chosen according to the available compute budget, to saturate the capacity of each GPU, without compromising the final loss, \textit{as long as it is less than $B_{\text{crit}}$}. 
\end{takeaways}

\paragraph{Critical batch size as function of model size.} In~\autoref{fig:ScritfromN} we plot the final $\exp{(\text{ELBO})}$ as a function of the iteration count $S$ and model's size $N$, for a constant token horizon $D$ of 13B tokens. We see that the critical iteration count $S_{\text{crit}}$ is \textbf{independent} of model size. This is compliant with the findings of~\citet{DBLP:journals/corr/abs-2505-13738}, albeit in the non-SDE case. The maximum per-GPU batch size typically decreases with model size until it reaches $1$, after which it requires more involved techniques (e.g., more fine-grained parallelization). Therefore, maintaining the same global batch size typically require more nodes as model size increases.    
\begin{takeaways}
Under SDE scaling, critical batch size is not impacted by model size. At equal token horizon, a batch size less than $B_{\text{crit}}$ for a smaller model is also safe for a bigger model.
\end{takeaways}
  
\paragraph{Critical batch-size as a function of the token horizon.} In~\autoref{fig:ScritfromN} we plot the final $\exp{(\text{ELBO})}$ as a function of the iteration count $S$ and model size $D$, for a constant model size $N$ of 320M parameters. We see that $S_{\text{crit}}$ grows with $D$, at \textit{sub-linear} speed, implying that the corresponding critical-batch size grows sub-linearly with the token horizon.  
\begin{takeaways}
The number of GPUs cannot be scaled proportionally with $D$ forever: there is a risk of reaching $B_{\text{crit}}$ when $D$ is sufficiently large. Beyond that point, either the marginal utility of every new token drops, or total wall-clock time increases. Over-trained models cannot be both FLOP-efficient and fast to train.
\end{takeaways}
\subsection{Optimal Drift--horizon Tradeoffs}
\begin{figure}[ht]
    \centering
    \begin{minipage}[t]{0.48\linewidth}
        \centering
        \includegraphics[width=\linewidth]{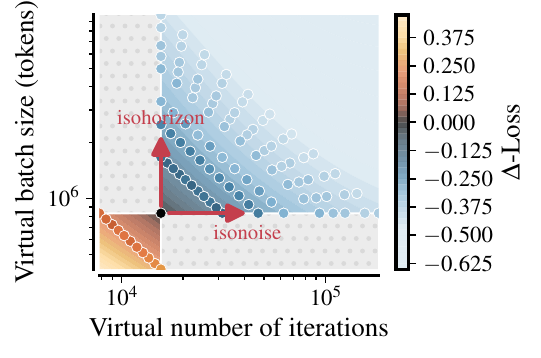}
        \caption{\textbf{Drift--horizon tradeoff~$\gamma$}. 
        When increasing token horizon $D$, the optimal choice lies between reducing the drift (increasing the virtual batch size) and increasing the SDE horizon (increasing the virtual number of iterations).}
        \label{fig:isonoiseisohorizon}
    \end{minipage}\hfill
    \begin{minipage}[t]{0.48\linewidth}
        \centering
        \includegraphics[width=\linewidth]{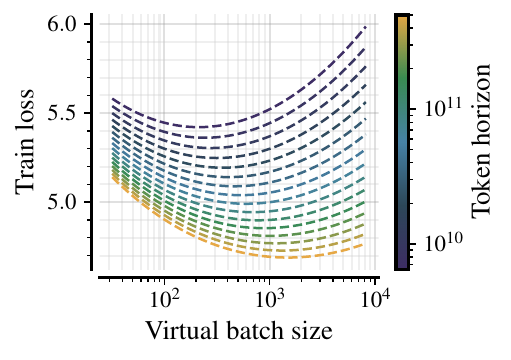}
        \caption{\textbf{Isotoken curves at various virtual batch sizes.} The optimal allocation between virtual batch size and virtual iteration count corresponds to a factor $\gamma^*\approx 0.44$. The bottom of the U-curve would be $B_{\text{opt}}$ in non-SDE parametrization.}
        \label{fig:isonoiseisohorizon_bis}
    \end{minipage}
\end{figure}

Normally, the physical batch size $B$ can be chosen freely to accommodate the number of nodes available, as long as $B\leq B_{\text{crit}}$. The SDE re-parametrization allows to re-map this to a \textit{virtual batch size} $\tilde{B}$ and a corresponding \textit{virtual number of iterations} $\tilde{S}$ such that $D=\tilde{B}\tilde{S}L$. They correspond to the behavior of the same model trained without SDE parametrization, and physical batch size $\tilde{B}$. This raises a question: since the physical batch size $B$ is chosen based on compute considerations only, how should we chose the virtual batch size $\tilde{B}$ when $D$ is scaled-up? To answer the question, we design an experiment in which they evolve jointly as:
\begin{equation*}
    \tilde{S}= G(D/L)^{1-\gamma}\quad\text{ and } \quad\tilde{B}=G^{-1}(D/L)^{\gamma},\qquad\text{ with }\qquad\gamma=\alpha/(\alpha + \beta)\quad\text{ and }\quad G = \left(\frac{\alpha A}{\beta B}\right)^{1/(\alpha+\beta)}.
\end{equation*}
The behavior $\gamma=0$ correspond to the default setup of the literature in training LLMs~\citep{DBLP:journals/corr/abs-2001-08361}, when more tokens simply correspond to more iterations, whereas the setup $\gamma=1$ is the choice made in the work of~\citet{mlodozeniec2025completed} that instead modulates the optimization hyperparameters (\autoref{sec:isonoisedef}). By sweeping over $[0,1]$ we decide how many of these extra-tokens are assigned to \textit{reducing the drift}, or to \textit{extending the horizon}. Results are given in~\autoref{fig:isonoiseisohorizon} and~\autoref{fig:isonoiseisohorizon_bis}. Surprisingly, neither setting used in the literature are optimal. By fitting a power law of the form $E+A\tilde{S}^{-\alpha}+B\tilde{B}^{-\beta}$ we find coefficients $\alpha=0.18,\beta=0.23$. Minimizing this parametric equation under the constraint $D=\tilde{B}\tilde{S}L$, we find that $\gamma^*=0.44$.  

\paragraph{Link with learning rate schedule.} For $\gamma < 1$, the effective learning rate rises with the physical batch size, leveraging the lower variance in gradient estimates for larger updates. Conversely, when $\gamma > 0$, a longer token horizon reduces the effective learning rate, permitting smaller incremental steps and deeper exploration of loss basins. Thus, $\gamma$ plays a role akin to a learning rate scheduler, determining the allocation of larger versus smaller updates. Consequently, the optimal $\gamma$ could shift if, for example, a warmup-stable-decay schedule~\citep{DBLP:conf/nips/HageleBKAWJ24,DBLP:conf/icml/SchaippHTS025} were used instead of the current cosine schedule.

\subsection{Scaling Laws for Tri-modal \gls{mdm}}
\label{sec:trimodal_scaling_laws}
\begin{figure*}
    \includegraphics[width=0.98\linewidth]{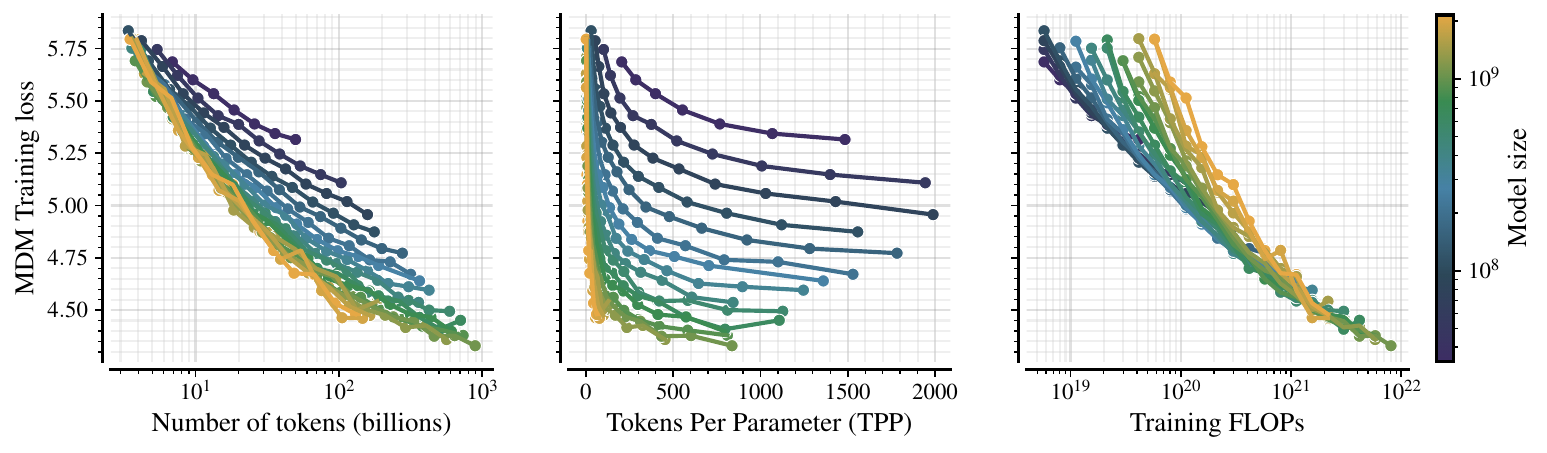}
    \caption{\textbf{Training curves for tri-modal \gls{mdm}.} We select 24 log-distributed isoFLOPS between 5e18 and 1e22 to cover the $(N, D)$ grid. The performance is dominated by the total compute budget $C$, following formula in appendix \textbf{H.1} of~\citet{DBLP:conf/icml/BusbridgeSWRLW25}. The loss is computed on an independent validation set with identical mixture weights.}
    \label{fig:trimodaltrainingcurve}
\end{figure*}

\begin{figure}[tbh]
    \centering
    \begin{minipage}[t]{0.48\linewidth}
        \centering
        \includegraphics[width=\linewidth]{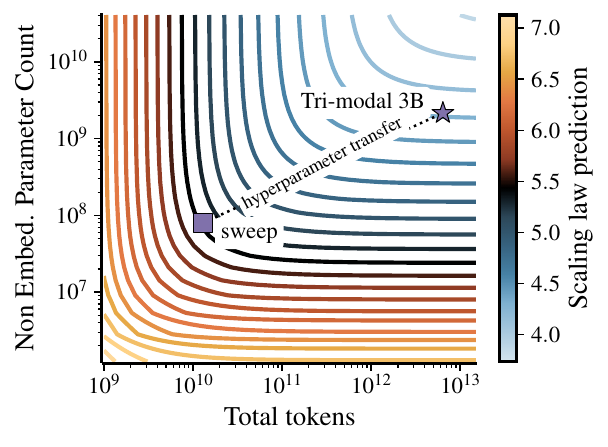}
        \caption{\textbf{Isoloss contours for tri-modal MDM}. Dashed line indicates direction in which the 0-shot hyperparameter transfer is done using CompleteP + SDE.}
        \label{fig:isocontour}
    \end{minipage}\hfill
    \begin{minipage}[t]{0.48\linewidth}
        \centering
        \includegraphics[width=\linewidth]{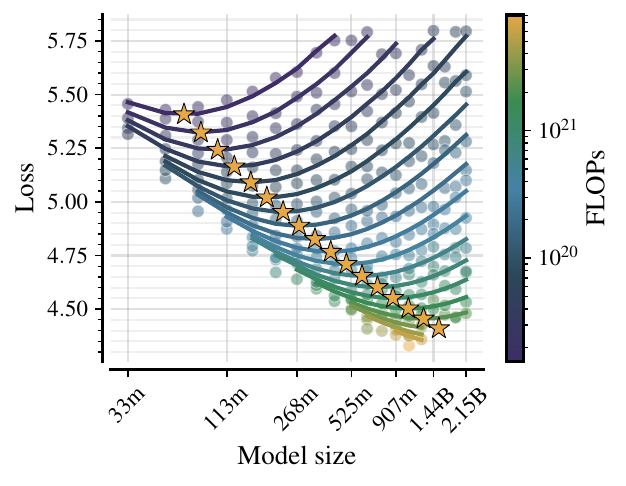}
        \caption{\textbf{IsoFLOPs for tri-modal MDMs.} Solid lines indicate scaling law predictions. Points represent measurements and $\star$ indicates the lowest loss achievable at each isoFLOP.}
        \label{fig:trimodalisoflops}
    \end{minipage}
\end{figure}

Scaling laws provide a prescriptive mechanism to decide the model size ($N$) and the number of tokens ($D$) in a compute optimal manner. They are obtained by fitting power laws from raw training curves, as in~\autoref{fig:trimodaltrainingcurve}. One of the core contributions of our work is in establishing this for tri-modal \gls{mdm}s that have been scaled under CompleteP with SDE re-parametrization. We train \emph{262 different tri-modal \gls{mdm} models} with Token Per Parameter (TPP) ratios between 1 and 2000. We sample the $(N,D)$ pairs along 24 different isoFLOPs logarithmically distributed between
$5 \times 10^{18}$ and $1 \times 10^{22}$. The model's size here does not account for embedding parameters~\citep{DBLP:journals/tmlr/PearceS24}, which strongly impacts smaller models as shown in~\autoref{fig:ratiorealfakeparams}. FLOPs per token are computed using the formula in Appendix \textbf{H.1} of~\citet{DBLP:conf/icml/BusbridgeSWRLW25}. Modality tokenizers are not taken into account for the total FLOP budget. Following~\citet{DBLP:journals/corr/abs-2507-09404}, we fit a power law using basin hopping and LBFGS, with 20 bootstrap samples with 90\%-10\% cross-validation, to ensure stability of coefficients estimation. We found that the additive form was insufficient to explain the measurements, and had to rely on the form introduced in the seminal work of~\citet{DBLP:journals/corr/abs-2001-08361}, with an additional $E$ term to avoid loss degeneracy in the $N,D\rightarrow\infty$ limit.
\begin{equation}
    L = E + \left(\frac{A}{N^{a/b}}+\frac{B}{D}\right)^b.
    \label{eq:kaplan}
\end{equation}
We report scaling coefficients of $a\approx0.14$ and  $b\approx0.17$ in~\autoref{fig:sdefit}, using the method of~\Cref{sec:extended_scaling_laws}. We measure an $R^2$ score of 99.3\% and an MRE of 0.5\%. %
We report isoloss contours in~\autoref{fig:isocontour} and isoFLOP curves in~\autoref{fig:trimodalisoflops}. Finally, we compute optimal number of tokens per parameter as:
\begin{equation}
    D^\star(N)=7754\cdot N^{\alpha}\text{ with }\alpha=a/b=0.84.
\end{equation}
We report the \textit{compute-optimal} token count $D^{\star}(N)$ as function of model size against other popular model families in~\autoref{fig:pareto-front-flops}. For all compute-optimal curves, we use the same methodology: we plot token horizon $D^{\star}$ as function of \textit{total parameter count}, and rely on approach 3 of Chinchilla with corrected coefficients from~\citet{DBLP:journals/corr/abs-2404-10102}. We find that a 3B model requires at least $480$B tokens, in sharp contrast with $60$B tokens reported by Chinchilla~\citep{DBLP:journals/corr/abs-2203-15556} for autoregressive language models. This gap is maintained at all realistically reachable model sizes. 

\paragraph{Analysis.} These scaling laws suggest that as $N$ grows, the TPP ratio $D^{\star}/N \propto N^{-0.16}$ decreases, i.e., \glspl{mdm} become asymptotically more data-efficient per parameter. This is in contrast to AR language models, for which the rule of thumb $D \propto 20N$ popularized by~\citet{DBLP:journals/corr/abs-2203-15556} typically holds. The value of $b\approx0.17$ is compatible with the one found in experiment of~\Cref{sec:b_opt}, on iteration count $S$. The optimal compute allocation between tokens and parameters is then given by:
\begin{equation}
    N^{\star}(C)\propto C^{0.55}\qquad\text{ and }\qquad D^{\star}(C)\propto C^{0.45}.
\end{equation}
Our coefficients are slightly higher for $N$ than they are for $D$, suggesting diminishing returns of additional data when increasing model size. However, this asymptotic trend is offset at practical scales by the large leading constant ($D^{\star} = 7754 \cdot N^{0.84}$): a 3B model still requires ${\sim}480$B tokens, far exceeding the ${\sim}60$B implied by Chinchilla for autoregressive models. The crossing with Quokka~\citep{DBLP:journals/corr/abs-2510-03280} compute-optimal curves happens around the 20B scale ($\approx 2$T tokens). Below that crossing point, at equal FLOPs, models from our tri-modal MDM family should be \textit{smaller} and \textit{trained for longer} than the ones of Quokka. Above that crossing point, the trend reverses.

\paragraph{Interpretation.} While $a$ and $b$ coefficients inform on the relative effectiveness of parameter and data scaling within the same family of methods, they are \textit{not} sufficient in isolation to conclude superiority of a method over another. For example, some models such as logistic regression can still exhibit favorable exponents in some regimes~\citep{lin2024scaling} despite low expressiveness. In general, the asymptotics are better characterized by the value of $E$, corresponding to the incompressible error rate: the intrinsic entropy of the dataset, plus an additional error term coming from the bias of the family of models considered. 
Moreover, the ELBO is informative within a diffusion family but can be misleading across families, since different forward processes induce different likelihood bounds~\citep{sahoo2026scaling}; similarly, the data-coefficient $b$ is sensitive to data composition, as repetitions reduce the effective token count and thus deflate scaling efficiency~\citep{muennighoff2023scaling}.

\begin{figure}
    \begin{minipage}[t]{0.48\linewidth}
        \centering
        \includegraphics[width=\linewidth]{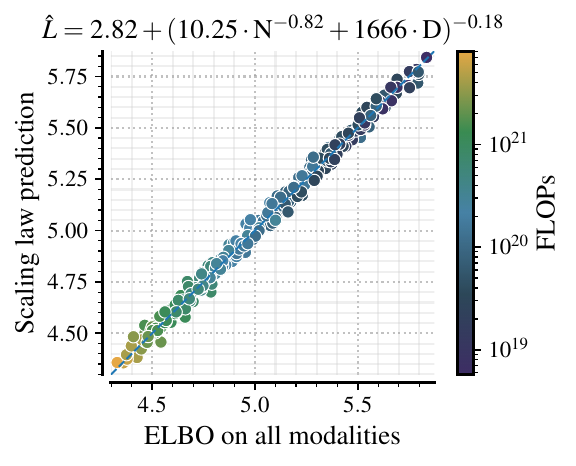}
        \caption{\textbf{Scaling law fit for tri-modal MDMs} using~\autoref{eq:kaplan} inspired by~\citet{DBLP:journals/corr/abs-2001-08361}. $R^2$ score of 99.3\% and an MRE of 0.5\%. $N$ and $D$ are expressed in billions units.}
        \label{fig:sdefit}
    \end{minipage}\hfill
    \begin{minipage}[t]{0.46\linewidth}
        \centering
        \includegraphics[width=\linewidth]{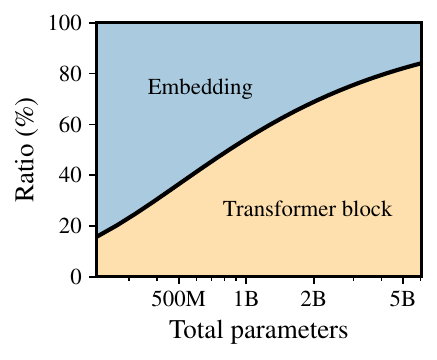}
        \caption{\textbf{Percentage of transformer-block parameters in the total parameter count}. 
        Tri-modality forces a larger vocabulary, yielding a ratio below 50\% for small models.}
    \label{fig:ratiorealfakeparams}
    \end{minipage}
\end{figure}

\section{Data}
\label{sec:data}

To further parallelize and simply our analysis we conduct all experiments (with the exception of the modality mixing ablation of~\Cref{sec:modality_mixing}) using a 33\% (pure text), 33\% (image-text), 33\% (audio-text) mixing ratio. Inside each modality, we use a predetermined reweighting of each mixture component, balancing quality and diversity. For all experiments, the token horizon $D$ is smaller than the total dataset size, ensuring that we operate in the regime of a single global epoch. Nonetheless, some small, high-quality sub-mixture components are repeated up to 4 times, well within estimated repetition tolerance~\citep{DBLP:journals/corr/abs-2507-15857}.

\paragraph{Text data.} Our text corpus is an aggregation of Nemotron-CC~\citep{su2025nemotron}; DCLM~\citep{li2024datacomp}; some subsets of The Pile~\citep{gao2020pile} including Wikipedia, HackerNews, Ubuntu IRC, Arxiv, DM-mathematics, Openwebtext; licensed StackOverflow data; and various high-quality synthetic data obtained from reasoning traces of Qwen-32B; and other licensed datasets. All corpora go through an additional level of cleaning and filtering to remove PII (Personally Identifiable Information) and other problematic content linked to licensing issues. Different splits of the same datasets with identical mixture ratios are used to compute the validation loss.  
\paragraph{Audio-text data.} Our audio training data consist of 2M hours of audio scraped from the web and transcribed by Whisper~\citep{DBLP:conf/icml/RadfordKXBMS23}. The data was extracted from a larger dataset that was PII filtered to remove private information and underwent a series of quality filters based on speech activity detection, dialogue detection, production quality, and production complexity~\citep{DBLP:journals/corr/abs-2502-05139}.
\paragraph{Image-text data.} Our image training data consists of an aggregation of multiple image and recaptioned text datasets such as CC3M~\citep{DBLP:conf/acl/SoricutDSG18}, CC12M~\citep{ DBLP:conf/cvpr/ChangpinyoSDS21}, COYO~\citep{kakaobrain2022coyo-700m}, and other licensed datasets from Manzano~\citep{DBLP:journals/corr/abs-2509-16197}. All samples are also PII filtered to remove private information.

\section{Results}
\label{sec:results}
First, we detail the benefits of this unified design at large scale in~\Cref{sec:flagship}. Then, we systematically ablate key design choices for our tri-modal discrete diffusion model. We evaluate the impact of different modality mixing ratios (\Cref{sec:modality_mixing}) and masking schedules during training (\autoref{sec:masking_schedule_ablation}). We examine inference-time hyperparameters for text-to-image generation and text-to-speech generation (\Cref{sec:inference_ablations}). Lastly, we explore the usage of anti-masking during training, which consists of augmenting the batch by generating two masked samples per sample, where one is the opposite of the other (\Cref{sec:anti_masking_ablation}). All ablations are conducted independently, starting from the same setup. %
\subsection{Unified 3B Tri-modal \gls{mdm}} 
\label{sec:flagship}
\begin{figure*}[ht]
\centering
\begin{minipage}[t]{0.40\textwidth}
  \vspace{0pt}
  \centering
  \includegraphics[width=\linewidth,trim=2mm 2mm 2mm 2mm,clip]{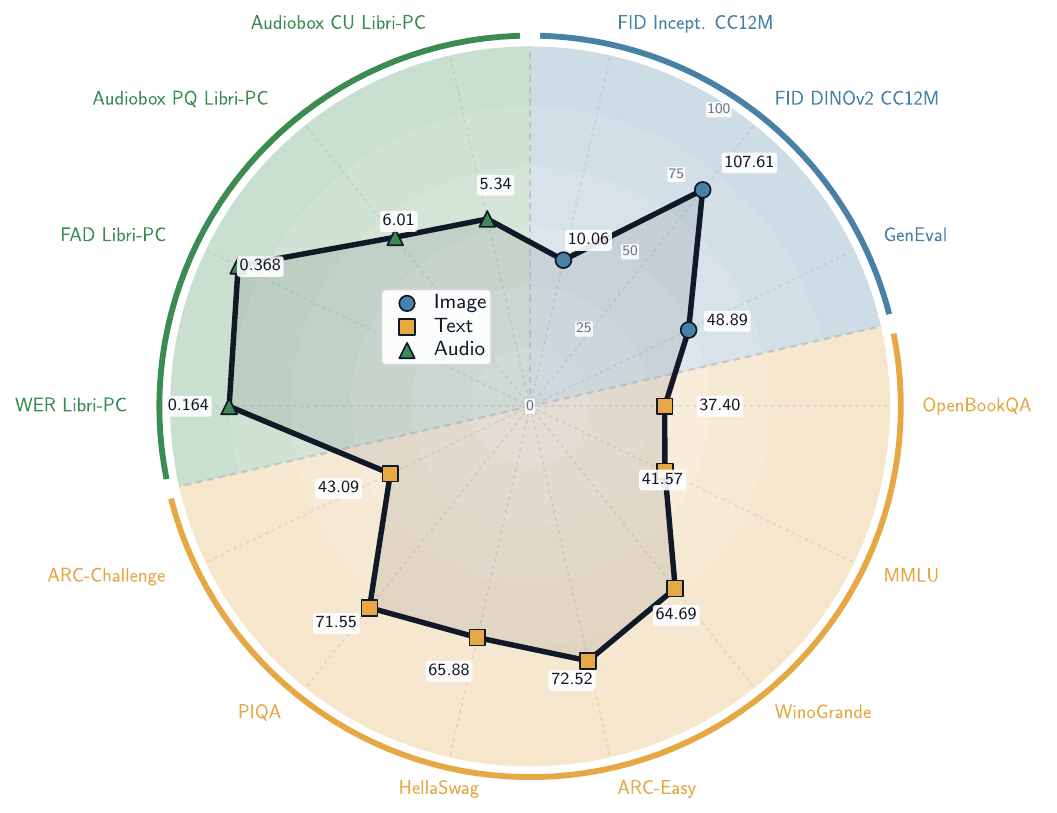}
\end{minipage}\hfill
\begin{minipage}[t]{0.58\textwidth}
  \vspace{0pt}
  \centering
  {\scriptsize
  \setlength{\tabcolsep}{3pt}
  \renewcommand{\arraystretch}{1.08}
  \begin{tabularx}{\linewidth}{@{}l >{\raggedright\arraybackslash}X@{}}
      \toprule
      \textbf{Modality} & \textbf{Dataset and Metric} \\
      \midrule
      \multicolumn{2}{@{}l}{\textbf{Image generation \& composition}} \\
      Train (eval seed) &
      FID-Inception 10.41, FID-DINOv2 112.12. \\
      CC12M & FID-Inception 10.06, FID-DINOv2 107.61. \\
      GenEval & Single Obj 93.12, Two Obj 63.38, Counting 33.44, Colors 64.89, Position 11.50, Color Attr. 27.00, Overall 48.89. \\
      \addlinespace[1mm]
      \multicolumn{2}{@{}l}{\textbf{Language (LM Harness)}} \\
      Harness & OpenBookQA 37.40, TruthfulQA MC2 40.76, BBH 24.97, MMLU 41.57, WinoGrande 64.69, ARC-Easy (norm) 72.52, HellaSwag (norm) 65.88, LogiQA2 (norm) 30.66, PIQA (norm) 71.55, 
      ARC-Challenge (norm) 43.09. \\
      \addlinespace[1mm]
      \multicolumn{2}{@{}l}{\textbf{Audio generation}} \\
      Train (eval seed) & FAD 0.218, WER 0.124, PQ 6.89, CU 6.20, CE 5.45, PC 1.89. \\
      LibriSpeech-PC & FAD 0.368, WER 0.164, PQ 6.01, CU 5.34, CE 5.07, PC 3.01. \\
      \bottomrule
  \end{tabularx}
  }
\end{minipage}
\caption{
  Tri-modal 3B overview across image, text, and audio.
  \textbf{Left:} radar summary (larger radius indicates better normalized score; vertex labels are raw values).
  Normalization is mixed by metric type: bounded metrics use natural bounds (GenEval/LM percentages to 100, WER to 1, Audiobox to 10), while unbounded Frechet metrics (FID-Incept, FID-DINOv2, FAD) use ECDF calibration from references, with lower-is-better inversion \(s=1-F(x)\), where \(F\) is the ECDF. \textbf{Right:} raw unnormalized metrics.
}
\label{fig:trimodal-3b-radar-table}
\end{figure*}
We evaluate our pretrained-only 3B tri-modal \gls{mdm} (see \autoref{tab:training_details} for the complete list of hyperparameters) on different settings for each specific modality. For text benchmarks we use LM evaluation harness (LM harness)~\citep{eval-harness}. For image benchmarks, we evaluate generation FID~\citep{DBLP:conf/nips/HeuselRUNH17}, using both DINOv2-L~\citep{DBLP:journals/tmlr/OquabDMVSKFHMEA24} and Inception-v3~\citep{DBLP:conf/cvpr/SzegedyVISW16} as feature extractors, computed on CC12M~\citep{DBLP:conf/cvpr/ChangpinyoSDS21} and the training data sampled with a different evaluation seed (hereafter \textit{train (eval seed)}), and the GenEval~\citep{DBLP:conf/nips/GhoshHS23} evaluation suite. For audio benchmarks, we evaluate text-to-speech generation conditioned on ground-truth prompts and durations from the train (eval seed) and LibriSpeech-PC~\citep{DBLP:conf/asru/MeisterNKBLG23} using generation FAD~\citep{DBLP:conf/interspeech/KilgourZRS19, DBLP:conf/icassp/GuiGBE24}, WER, and Audiobox Aesthetics~\citep{DBLP:journals/corr/abs-2502-05139} scores measuring four perceptual dimensions: Production Quality (PQ), Content Usefulness (CU), Content Enjoyment (CE), and Production Complexity (PC), as metrics. We present the broken down results for the individual modalities in the radar plot and table in \autoref{fig:trimodal-3b-radar-table}.

\subsection{Modality Mixing Ratios}
\label{sec:modality_mixing}

\begin{figure}[hbt]
    \begin{minipage}[t]{0.32\linewidth}
        \centering
        \includegraphics[width=\linewidth]{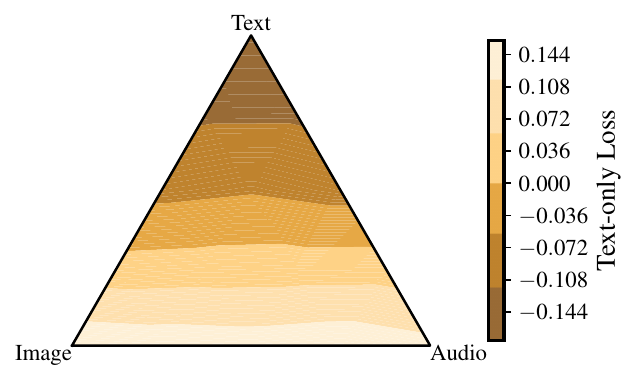}
    \end{minipage}
    \begin{minipage}[t]{0.32\linewidth}
        \centering
        \includegraphics[width=\linewidth]{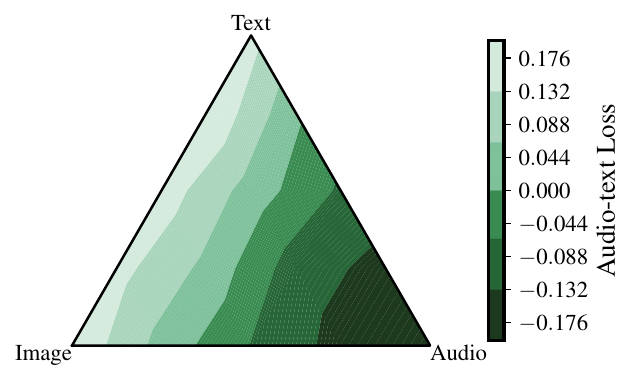}
    \end{minipage}
    \begin{minipage}[t]{0.32\linewidth}
        \centering
        \includegraphics[width=\linewidth]{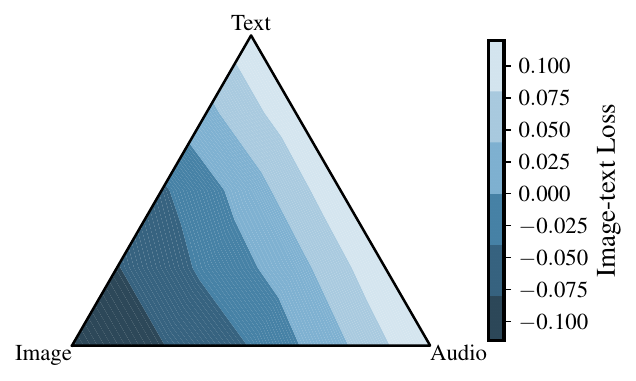}
    \end{minipage}    
    \caption{\textbf{Loss contours for tri-modal mixture coefficients}, taking $[1/3, 1/3, 1/3]$ as the reference point for the 0-level contour. We do not observe synergies between modalities at that model and data scale: they all compete for capacity and tokens.}
    \label{fig:mixtures}
\end{figure}
Understanding how to combine data is of critical importance for multimodal models. We take an empirical stride towards quantifying this by carefully constructing an experiment where we vary the global modality mixing ratios, $\{w_{\text{text}}, w_{\text{image}}, w_{\text{audio}}\}$, which respectively control the amount of text, image-text and audio-text data that is present within the pretraining mixture. We launch a total of 15 experiments with a model of size 320M (including. 80M non-embedding ones) and 13B tokens. We set a minimum of $20$\% per modality to avoid degeneracy with out-of-distribution modalities. Results are shown in~\autoref{fig:mixtures}. Unsurprisingly, we find that the loss decreases as the corresponding mixture weight increases. However, we do not witness synergies between modalities \emph{at these scales}: the average loss on a modality is independent on the relative ratio of the two other modalities. Therefore, the default choice of $[1/3, 1/3, 1/3]$ appears reasonable. While there exists prescriptive mechanisms to determine mixing ratios, they either fail to tackle multimodal models~\citep{chen2026olmix} or to account for data-repetition~\citep{DBLP:journals/corr/abs-2507-09404}.  
\subsection{Best Generation Hyperparameters}
\label{sec:inference_ablations}
\begin{figure}[ht]
    \centering
    \includegraphics[width=0.74\textwidth]{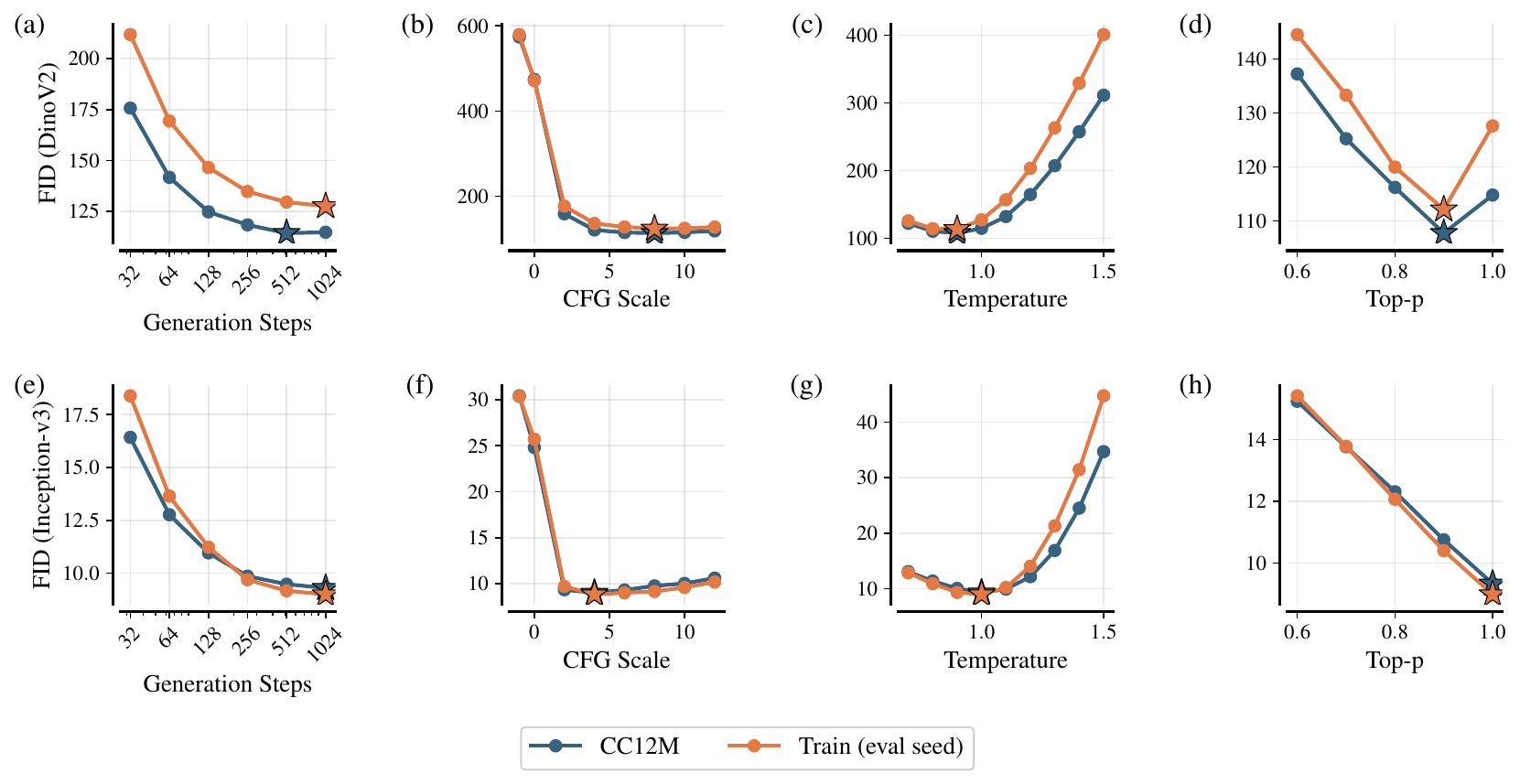}
    \caption{\textbf{Text-to-image hyperparameter ablations.}
    We generate images from text prompts and compute FID against reference images on two datasets: CC12M in blue and train (eval seed) in orange. Top row (a--d) shows FID computed using DINOv2-L features, bottom row (e--h) shows FID using Inception-v3 features. Panels (a,e) vary generation steps; (b,f) vary CFG scale; (c,g) vary temperature; (d,h) vary top-$p$. Stars indicate optimal values for each metric-dataset combination.}
    \label{fig:image_ablations}
\end{figure}
\begin{figure}[ht]
    \centering
    \includegraphics[width=\textwidth]{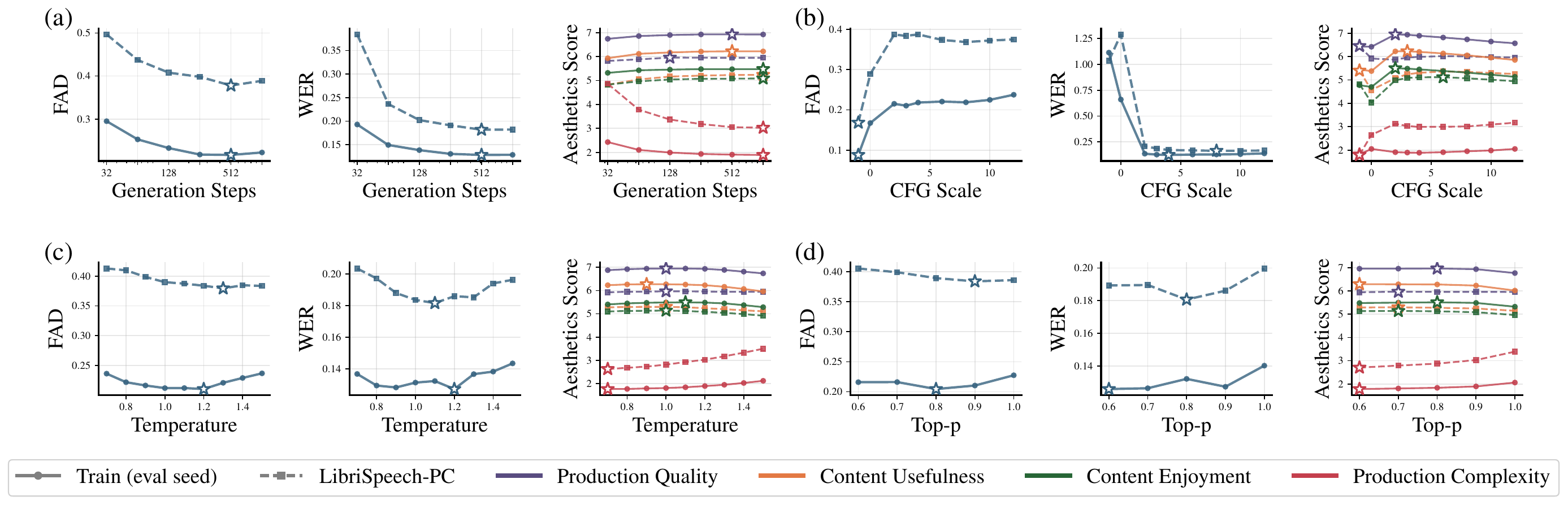}   
    \caption{\textbf{Text-to-speech hyperparameter ablations.}
  We evaluate audio generation from text prompts on two datasets (train set, LibriSpeech-PC) using ground-truth durations, with three metrics: FAD, WER, and Audiobox Aesthetics. Top row shows (a) steps sweep and (b) CFG sweep. Bottom row shows (c) temperature sweep and (d) top-$p$ sweep. $\star$ indicate the best hyperparameter for each metric. For Production Complexity, lower values indicate simpler audio and are considered as preferable.}
    \label{fig:audio_gt_length_ablations}
\end{figure}

We evaluate the impact of four different inference hyperparameters on generation quality: classifier-free guidance (CFG) scale, temperature, top-$p$ sampling, and number of generation steps. 
Unconditional generation for CFG is achieved by setting the text prompt to a fully masked state.
\paragraph{Text-to-image generation.} We generate 10,000 images conditioned on text prompts from CC12M and train (eval seed), using as default configuration of CFG=6.0, temperature=1.0, top-p=1.0, and 1024 generation steps, while ablating one hyperparameter at a time. As shown in \autoref{fig:image_ablations}, FID improves as the number of steps increases but with diminishing returns at higher step counts. For CFG, temperature, and top-p, optimal FID is achieved at intermediate values, with the exception of top-p which shows preference for higher values.
\paragraph{Text-to-speech generation.} We evaluate audio generation based on text prompts from our audio-text train (eval seed) dataset and LibriSpeech-PC. Here, we select 10,000 samples from each dataset and filter to retain only samples with duration $\leq 30$ seconds, retaining approximately 70\% of train (eval seed) samples and 99\% of LibriSpeech-PC samples. This filtering ensures consistent evaluation, as audio samples were truncated to a maximum token budget during training. Then, we use ground-truth durations for variable-length generation, allowing us to focus on the effect of sampling hyperparameters independent of duration prediction. Lastly, we ablated one hyperparameter at a time while relying on a default configuration of CFG=3.0, temperature=1.2, top-p=0.9, and 1000 generation steps.
As shown in \autoref{fig:audio_gt_length_ablations}, quality improves with more generation steps but with diminishing returns at higher step counts, similarly to image generation. CFG scale shows an interesting trade-off where increasing CFG strengthens text conditioning, improving transcription accuracy (WER), but degrades audio fidelity as captured by FAD. Stars indicate optimal hyperparameter values for each metric, where lower values are preferred for FAD, WER, and Production Complexity (simpler audio)~\citep{DBLP:journals/corr/abs-2502-05139}, while higher values are preferred for the other aesthetics metrics. CFG, temperature, and top-p exhibit varying optimal values across different metrics. Trends broadly stay consistent between the two datasets, demonstrating good generalization to the external LibriSpeech-PC dataset.

\subsection{Anti-Masking}
\label{sec:anti_masking_ablation}
The stochastic nature of the MDM training strategy is known to induce high variance~\citep{DBLP:conf/icml/RutteFDOS025}. To mitigate this, recent work introduced anti-masking~\citep{DBLP:journals/corr/abs-2511-18159,DBLP:journals/corr/abs-2506-20639}, which stabilizes training by applying decorrelated masks to each batch input. More specifically, one samples a standard mask and subsequently applies its negation to the same input, resulting in two masked versions of the same input sample. While this reduces variance in batch gradient estimates~\citep{DBLP:journals/corr/abs-2511-18159,DBLP:journals/corr/abs-2506-20639}, it doubles the computational cost of training as each sample in a batch is masked twice. We therefore compare models under a unique fixed token horizon $D$. To ensure compute matching, the baselines are trained with regular masking for two epochs, while the anti-mask variations are trained for a single epoch. To exemplify this setup, consider a dataset where $D$ consists of $8$ samples, and a batch size of $4$ is used. The anti-masking model would process the following sequence of batches: $[1, 1^*, 2, 2^*]$, $[3, 3^*, 4, 4^*]$, $[5, 5^*, 6, 6^*]$, $[7, 7^*, 8, 8^*]$, where $*$ denotes the repeated sample with complementary masking patterns. In contrast, standard training would iterate through the unique samples twice: $[1, 2, 3, 4]$, $[5, 6, 7, 8]$, $[1, 2, 3, 4]$, $[5, 6, 7, 8]$. For simplicity, we show the same order of samples across two epochs, but in practice samples are re-ordered.

We then conduct ablation studies on both text-only and multimodal architectures. The text models are approximately 7B parameters and are trained with a budget of 100 unique tokens per parameter. The multimodal models, with roughly 1.3B parameters, are trained on a horizon of 50 unique tokens per parameter. We evaluate the impact of anti-masking by comparing these models against their standard baselines, using a subset of the LM Harness for text tasks and assessing generation quality for audio and images in the multimodal setting.

\begin{table}[h]
    \centering
    \caption{\textbf{Multimodal ablation results comparing standard MDM training versus anti-masking.}}
    \label{table:multimodal_antimask_ablation}
    \scalebox{0.90}{
    \begin{tabular}{l cc cc cc}
        \toprule
        \multirow{2}{*}{\textbf{Model}} & \multicolumn{2}{c}{\textbf{FID (Inception)} $\downarrow$} & \multicolumn{2}{c}{\textbf{FID (DINOv2)} $\downarrow$} & \multicolumn{2}{c}{\textbf{FAD} $\downarrow$} \\
        \cmidrule(lr){2-3} \cmidrule(lr){4-5} \cmidrule(lr){6-7}
         & Train Data & CC12M & Train Data & CC12M & Train Data & LibriSpeech \\
        \midrule
        Base & 18.69 & 26.77 & 306.19 & 395.73 & 0.24 & 0.79 \\
        Anti-mask & \textbf{17.81} & \textbf{21.04} & \textbf{302.61} & \textbf{361.00} & \textbf{0.22} & \textbf{0.55} \\
        \bottomrule
    \end{tabular}
    }
\end{table}

Results for the multimodal experiments are presented in \autoref{table:multimodal_antimask_ablation}, reporting FID for images, and FAD for audio. We observe that anti-masking yields a positive impact on performance across modalities, with the most significant gains observed in audio generation quality. Results for the text-only models are provided in \autoref{tab:text_harness_results}. We see consistent improvements across most the tasks.

\begin{table}[!hbt]
    \centering
    \caption{\textbf{Anti-masking results on the evaluation harness.} We report mean accuracy $\pm$ standard deviation.}
    \label{tab:text_harness_results}
    \resizebox{0.98\textwidth}{!}{
    \begin{tabular}{l c c c c c c c c c c}
        \toprule
        & OpenBookQA & TruthfulQA & BBH & MMLU & Winogrande & ARC-Easy & HellaSwag & LogiQA2 & PIQA & ARC-Challenge \\
        \midrule
        \textbf{Base} 
        & 32.80 $\pm$ 2.10 
        & \textbf{43.12} $\pm$ 1.45 
        & 21.52 $\pm$ 0.46 
        & 30.20 $\pm$ 0.39 
        & 52.01 $\pm$ 1.40 
        & 63.72 $\pm$ 0.99 
        & 52.20 $\pm$ 0.50 
        & 24.75 $\pm$ 1.09 
        & 67.19 $\pm$ 1.10 
        & 31.66 $\pm$ 1.36 \\
        \textbf{Anti-mask} 
        & \textbf{33.00} $\pm$ 2.10 
        & 36.97 $\pm$ 1.41 
        & \textbf{27.05} $\pm$ 0.50 
        & \textbf{32.86} $\pm$ 0.39 
        & \textbf{54.06} $\pm$ 1.40 
        & \textbf{66.04} $\pm$ 0.97 
        & \textbf{55.15} $\pm$ 0.50 
        & \textbf{26.27} $\pm$ 1.11 
        & \textbf{67.41} $\pm$ 1.09 
        & \textbf{34.90} $\pm$ 1.39 \\
        \bottomrule
    \end{tabular}
    }
\end{table}

\begin{takeaways}
Anti-masking is a simple yet effective approach to improve MDM pre-training in the multiple-epochs setup. Crucially, it improves benchmark performance without incurring additional computational cost.
\end{takeaways}

\section{Conclusion}
\label{sec:conclusion}
This work reframes multimodal generation as order agnostic iterative refinement by extending masked discrete diffusion from language to a unified tri-modal setting, where text, images, and audio share a single token stream and a single transformer backbone, enabling flexible conditioning (captioning, text to image, \gls{asr}, \gls{tts}) without modality specific heads or bespoke factorizations. Beyond demonstrating feasibility, we chart the practical design space that governs stability and efficiency at scale: we show how SDE-based reparameterization %
reduce expensive tuning, we derive empirical scaling behavior to guide compute optimal training, and we surface a strongly modality dependent inference landscape where sampling hyperparameters sampling (guidance, temperature, steps) must be chosen differently for different modalities. Lastly, targeted training interventions such as anti-masking yield consistent improvements under compute matched comparisons. %

\clearpage
\bibliography{biblio}
\bibliographystyle{unsrtnat}

\clearpage
\onecolumn
\appendix
\setcounter{tocdepth}{3}
\setcounter{secnumdepth}{3}
\appendixpage

\startcontents[sections]

\printcontents[sections]{l}{1}{}

\clearpage

\section{General formulation of weighting and the masking process}
\label{app:weighting}
In this section, we expand the \gls{mdm} forward process explained in \Cref{sec:method} and make it consistent with the previously introduced notation in the literature, particularly following \citet{DBLP:conf/nips/ShiHWDT24}.

In the general case of \gls{mdm}, we progressively corrupt the original data $s_0 \in \gV^{L^\star}$ into a masked version $s_t$ over $T$ discrete time steps. The forward process defines a Markov chain that transforms an original data $s_0$ into a corrupted version $s_t$ at time step $t \in [T]$. This process is governed by a sequence of transition matrices for each position $i$ denoted by $Q_t^i \in \R^{V \times V}$. At each step $t$, a token $s_{t-1}^i$ is transformed into $s_{t}^i$ according to the probability $q(s_{t}^i \mid s_{t-1}^i)$. We define the masking probability at time $t$ as $\beta_t \in [0, 1]$. Consequently, the probability of a token \textit{not} being masked is $\alpha_t = 1 - \beta_t$.

The single-step transition matrix $Q_t^i$ is typically defined as:
\begin{align*}
    Q_t^i(v \mid r) = q(s_t^i=v \mid s_{t-1}^i=r) = \begin{cases}
    1 - \beta_t & \text{if }  r\neq \text{MASK}_{m(i)}\text{ and } v=r \\ 
    \beta_t & \text{if } r\neq \text{MASK}_{m(i)}  \text{ and } v=\text{MASK}_{m(i)} \\
    1 & \text{if } r=v=\text{MASK}_{m(i)} \\
    0 & \text{otherwise} \end{cases}
\end{align*}

This formulation implies that a token either remains unchanged or is replaced by the $\text{MASK}_{m(i)}$ token. More general formulations might allow transitions to any other token with a small probability.

The \emph{cumulative} transition probability from $s_0$ to $s_t$ is crucial for training and is given by the product of individual transition matrices: $Q_t^i(v \mid r) = q(s_{t}^i=v \mid s_{0}^i=r)$. This can be simplified by defining $\bar{\alpha}_t = \prod_{\tau=1}^t \alpha_\tau = \prod_{\tau=1}^t (1 - \beta_\tau)$. The probability of a token $s_0^i$ remaining unchanged until time $t$ is $\bar{\alpha}_t$. Conversely, the probability of it having been masked at least once and therefore being a $\text{MASK}_{m(i)}$ token at time $t$ is $1 - \bar{\alpha}_t$. Therefore, the marginal distribution of $s_t$ given $s_0$ for a single token $i$ is:
\begin{align*}
    q(s_{t}^i=v \mid s_{0}^i=r) = \begin{cases} \bar{\alpha}_t & \text{if } v=r \\ 1 - \bar{\alpha}_t & \text{if } v=\text{MASK}_{m(i)} \\ 0 & \text{otherwise} \end{cases}\,, \qquad q(s_t \mid s_0)
    = \prod_{i=1}^{L^\star} q(s_t^i \mid s^i_0).
\end{align*}

This distribution $q(s_t \mid s_0)$ allows for direct sampling of $s_t$ from $s_0$ at any time step $t$.

\subsection{Connection of Loss Weighting and Cumulative Corruption}
\label{app:w_alpha_connection}
The weighting function $w(t)$ plays a critical role in balancing the contribution of different time steps to the total loss, with the choice of $w(t)$ being intimately connected to the masking schedule defined by $\alpha_t$ or equivalently $\beta_t$.

Recall that $\bar{\alpha}_t$ represents the cumulative probability that a token \emph{has not been masked} up to time $t$. Conversely, $1 - \bar{\alpha}_t$ is the probability that a token \emph{has been masked} by time $t$. This results in two opposite scenarios:
\begin{itemize}
    \item \textbf{Early time steps} (small $t$, $\bar{\alpha}_t \approx 1$): Few tokens are masked. Predicting the original $s_0$ for these few masked tokens is relatively easy, as most of the context (unmasked tokens) is available.
    \item \textbf{Late time steps} (large $t$, $\bar{\alpha}_t \approx 0$): Most tokens are masked. Predicting the original $s_0$ becomes very challenging, requiring the model to infer from minimal context.
\end{itemize}

A common motivation for weighting comes from \gls{elbo} in continuous diffusion, which often leads to weighting terms that compensate for varying noise levels. Based on \Cref{app:weighting}, the forward marginal distribution is
\begin{align*}
    q(s_t \mid s_0)
= \prod_{i = 1}^{L^\star}
\bar{\alpha}_t\,\delta_{s_0^i}(s_t^i)
+
\big(1-\bar{\alpha}_t\big)\,\delta_{\text{MASK}_{m(i)}}(s_t^i)\,.
\end{align*}

Training maximizes the \gls{elbo}, which decomposes into timestep-wise \gls{kld}:
\begin{align*}
\Ls
=
\sum_t
\E_{q}
\Big[
\mathrm{KL}\big(
q(s_{t-1}\mid s_t,s_0)
\;\|\;
p_\theta(s_{t-1}\mid s_t)
\big)
\Big]\,.
\end{align*}

In \gls{mdm}, each KL term is non-zero only when $s_t^i=\text{MASK}_{m(i)}$. Conditioning on this event, the posterior
$q(s_{t-1}^i\mid s_t^i=\text{MASK}_{m(i)},s_0)$
is a categorical distribution whose parameters depend on the incremental masking rate. More precisely, let us denote $\pi_t$ the probability that masking occurred at time $t$ rather than earlier, so that
\begin{align*}
    q(s_{t-1}^i\mid s_t^i=\text{MASK}_{m(i)},s_0) = \pi_t \delta_{s_0^i}(s_{t-1}^i) + (1-\pi_t) \delta_{\text{MASK}_{m(i)}}(s_{t-1}^i)\,.
\end{align*}
Using Bayes' rule, we have:
\begin{align*}
    \pi_t = \frac{\mathbb{P}(s_{t-1}^i = s_0^i)\mathbb{P}(s_{t}^i = \text{MASK}_{m(i)} \mid s_{t-1}^i = s_0^i)}{\mathbb{P}(s_t^i = \text{MASK}_{m(i)})}\,,
\end{align*}
where $\mathbb{P}(s_{t-1}^i = s_0^i) = \bar{\alpha}_{t-1}$, $\mathbb{P}(s_{t}^i = \text{MASK}_{m(i)} \mid s_{t-1}^i = s_0^i) = 1-\alpha_t$, and $\mathbb{P}(s_t^i = \text{MASK}_{m(i)}) = 1 - \bar{\alpha}_t$. Therefore, we have:
\begin{align*}
    \pi_t = \frac{\bar{\alpha}_{t-1} (1-\alpha_t)}{1 - \bar{\alpha}_t} = \frac{\bar{\alpha}_{t-1} - \bar{\alpha}_{t}}{1 - \bar{\alpha}_{t}}\,.
\end{align*}
In the continuous-time limit, $\pi_t = \frac{-\bar{\alpha}_t'}{1 - \bar{\alpha}_{t}}$. As a result, the \gls{elbo} is equivalent up to constants to minimizing the objective
\begin{align*}
    \E_{t \sim \text{U}(0, 1)}
\left[
w(t)\;
\sum_{i = 1}^{L^\star}\E_{q(s_t^i \mid s_0^i)}
\big[
\ell_i(\theta, s_0)
\;\big|\;
s_t^i=\text{MASK}_{m(i)}
\big]
\right],
\qquad
w(t)
=
\frac{\bar{\alpha}'_t}{1-\bar{\alpha}_t} .
\end{align*}
This weighting ensures that each timestep contributes proportionally to the rate at which information about $s_0$ is destroyed by masking.

\subsection{Unbiasedness of the Importance Weighting}
At a fixed timestep $t$, each token position is independently masked with probability $t$, so that
$\mathbf{1}\{ i \in \gI_t \} \sim \mathrm{Bernoulli}(t)$.
Averaging the reconstruction loss only over masked positions therefore corresponds to random subsampling of tokens.

Without reweighting, the expected contribution of token $i$ to the loss in linear scheduling is
\begin{align*}
    \E\!\left[
\mathbf{1}\{ i \in \gI_t \}\,\ell_i(\theta, s)
\right]
=
t\,\ell_i(\theta, s) \,,
\end{align*}
which underweights tokens at small $t$ and biases the objective toward high-noise regimes. Multiplying by $1/t$ yields an unbiased estimator:
\begin{align*}
    \E\!\left[
\frac{1}{t}\,
\mathbf{1}\{ i \in \gI_t \}\,\ell_i(\theta, s)
\right]
=
\ell_i(\theta, s)\,.
\end{align*}
Therefore, the $1/t$ factor corrects for the subsampling induced by masking, ensuring that each token contributes equally in expectation across timesteps. This corresponds to inverse-probability weighting and is analogous to the time-dependent weighting used in denoising score matching objectives for diffusion models, where losses are rescaled to normalize signal-to-noise ratio across noise levels.

\section{Tokenizer Ablations}
\label{app:tokenizer_ablations}

\subsection{Audio Tokenizer Ablations}
\label{app:audio_tokenizer_ablations}

\begin{table}[ht]
\centering
\resizebox{0.95\textwidth}{!}{
\setlength{\tabcolsep}{4pt}
\begin{tabular}{lccccccc}
\toprule
\textbf{Model} & \textbf{\# Codebooks} & \textbf{PESQ} $\uparrow$ & \textbf{Content Enjoy} $\uparrow$ & \textbf{Content Useful.} $\uparrow$ & \textbf{Prod. Complex.} $\uparrow$ & \textbf{Prob. Quality} $\uparrow$ & \textbf{Down. Factor} $\uparrow$ \\
\midrule
\multirow{2}{*}{Higgs pretrained} & 8 & 3.168 & 5.670 & 6.192 & 1.559 & 6.510 & 960 \\
                                  & 4 & \textbf{2.544} & \textbf{5.577} & \textbf{6.087} & \textbf{1.561} & \textbf{6.390} & \textbf{960} \\
\multirow{2}{*}{DAC pretrained}   & 9 & 3.658 & 5.555 & 5.994 & 1.567 & 6.255 & 512 \\
                                  & 4 & 2.396 & 5.069 & 5.415 & 1.603 & 5.597 & 512 \\
\multirow{2}{*}{DAC retrained}    & 9 & 2.969 & 5.691 & 6.234 & 1.556 & 6.550 & 1024 \\
                                  & 4 & \textbf{2.433} & \textbf{5.585} & \textbf{6.131} & \textbf{1.566} & \textbf{6.434} & \textbf{1024} \\
\bottomrule
\end{tabular}
}
\caption{\textbf{Reconstruction metrics for different audio tokenizers.}}
\label{tab:tokenizer-results}
\end{table}

The audio tokenizer determines the audio token rate and, therefore, the sequence length corresponding to
the audio stream. Since we use a fixed context length ($L^{\star}=3256$) and clips with at most 30 seconds, we need
a low-rate codec that still preserves perceptual quality. Here, we compare several RVQ-based codecs: a pretrained 24\,kHz DAC~\citep{DBLP:conf/nips/KumarSLKK23}, a pretrained Higgs Audio v2~\citep{higgsaudio2025}, and a DAC-style tokenizer trained on the same data as the main model. In \autoref{tab:tokenizer-results}, we present the reconstruction evaluation on LibriTTS-clean using PESQ~\citep{DBLP:conf/icassp/RixBHH01} and Audiobox. Among the options, only two configurations fit our token budget for 30\,s audio, which we display in bold.
Based on these results, we chose the Higgs Audio v2 tokenizer with 4 codebooks decoding as the default audio tokenizer as it provides a strong and convenient rate--distortion trade-off.
As expected, increasing the number of codebooks improves reconstruction but quickly becomes impractical under our
sequence-length constraint. Consistent with prior observations for RVQ codecs~\citep{DBLP:conf/nips/KumarSLKK23},
we also find that training with more codebooks and decoding with fewer can preserve perceptual quality substantially
better than training the low-rate model directly.

\begin{table}[htbp]
\centering
\resizebox{0.95\textwidth}{!}{
\begin{tabular}{lccccc}
\toprule
\textbf{Model} & \textbf{Type} & \textbf{Latent / Code Dim} & \textbf{Vocab Size} & \textbf{Tokens Per Image} & \textbf{rFID} $\downarrow$ \\
\midrule
\multicolumn{6}{c}{\textbf{CC12M}} \\
\hline
cosmos-ci8x8~\citep{DBLP:journals/corr/abs-2501-03575} & Continuous & 16 & - & 1024 & 1.37 \\
cosmos-di16x16~\citep{DBLP:journals/corr/abs-2501-03575} & FSQ & 6 & 65536 & 256 & 3.50 \\
cosmos-di8x8-360p~\citep{DBLP:journals/corr/abs-2501-03575} & FSQ & 6 & 65536 & 1024 & 1.80 \\
ibq-262144~\citep{shi2025scalableimagetokenizationindex} & IBQ & 256 & 262144 & 256 & 0.89 \\
\textbf{movqgan-270m}~\citep{sber_movqgan}  & \textbf{MoVQ} & \textbf{256} & \textbf{16384} & \textbf{1024} & \textbf{0.50} \\
openmagvitv2~\citep{luo2025openmagvit2opensourceprojectdemocratizing} & LFQ & 18 & 262144 & 256 & 0.82 \\
unitok~\citep{unitok} & MCQ & 64 & 4096 (x8)* & 256 (x8) & 0.54 \\
\hline
\multicolumn{6}{c}{\textbf{ImageNet}} \\
\hline
cosmos-ci8x8~\citep{DBLP:journals/corr/abs-2501-03575} & Continuous & 16 & - & 1024 & 1.02 \\
cosmos-di16x16~\citep{DBLP:journals/corr/abs-2501-03575} & FSQ & 6 & 65536 & 256 & 4.38 \\
cosmos-di8x8-360p~\citep{DBLP:journals/corr/abs-2501-03575} & FSQ & 6 & 65536 & 1024 & 0.95 \\
ibq-262144~\citep{shi2025scalableimagetokenizationindex} & IBQ & 256 & 262144 & 256 & 1.55 \\
movqgan-270m~\citep{sber_movqgan} & MoVQ & 256 & 16384 & 1024 & 0.55 \\
openmagvitv2~\citep{luo2025openmagvit2opensourceprojectdemocratizing} & LFQ & 18 & 262144 & 256 & 1.67 \\
\textbf{unitok}~\citep{unitok} & \textbf{MCQ} & \textbf{64} & \textbf{4096 (x8)*} & \textbf{256 (x8)} & \textbf{0.36} \\
\hline
\multicolumn{6}{c}{\textbf{ImageNet - 512$\times$512}} \\
\hline
\textbf{cosmos-ci8x8}~\citep{DBLP:journals/corr/abs-2501-03575} & \textbf{Continuous} & \textbf{16} & - & \textbf{4096} & \textbf{0.07} \\
cosmos-di16x16~\citep{DBLP:journals/corr/abs-2501-03575} & FSQ & 6 & 65536 & 1024 & 1.33 \\
cosmos-di8x8-360p~\citep{DBLP:journals/corr/abs-2501-03575} & FSQ & 6 & 65536 & 4096 & 0.51 \\
ibq-262144~\citep{shi2025scalableimagetokenizationindex} & IBQ & 256 & 262144 & 1024 & 0.50 \\
\textbf{movqgan-270m}~\citep{sber_movqgan} & \textbf{MoVQ} & \textbf{256} & \textbf{16384} & \textbf{4096} & \textbf{0.17} \\
openmagvitv2~\citep{luo2025openmagvit2opensourceprojectdemocratizing} & LFQ & 18 & 262144 & 1024 & 0.53 \\
unitok~\citep{unitok} & MCQ & 64 & 4096 (x8)* & 1024 (x8) & 0.23 \\
\bottomrule
\end{tabular}
}
\caption{\textbf{Reconstruction FID for different image tokenizers.} 
ImageNet consists of 50k validation examples and CC12M~\citep{DBLP:conf/cvpr/ChangpinyoSDS21} consists of 50k samples from the full dataset. Best FID per section is shown in \textbf{bold}. When a continuous model achieves the best FID, the best discrete model is also bolded. *UniTok uses 8 categorical predictions of size 4096 (x8 underlying codes that are merged into a single token).}
\label{tab:tokenizer_model_results}
\end{table}

\subsection{Image Tokenizer Ablations}
\label{app:image_tokenizer_ablations}

The same sequence length constrains apply to the image tokenizer. We want a discrete image tokenizer that maps an image to as few tokens as possible while maintaining good image representation. Here, we compare discrete versions of Cosmos~\citep{DBLP:journals/corr/abs-2501-03575}, IBQ~\citep{shi2025scalableimagetokenizationindex}, OpenMagVIT~\citep{luo2025openmagvit2opensourceprojectdemocratizing}, Unitok~\citep{unitok}, and MoVQGAN~\citep{sber_movqgan}. We evaluate the reconstruction FID on ImageNet at 256 and 512 resolution as well as CC12M~\citep{DBLP:conf/cvpr/ChangpinyoSDS21} and present these results in \autoref{tab:tokenizer_model_results}. Based on these results, we chose MoVQGAN as the default image tokenizer as it provided a good balance between high reconstruction performance, high compression and small vocabulary size.

\section{Training details}
\label{app:training_details}
\subsection{Optimal Global Hyper-Parameter Search}
\label{sec:opt_globals}
\begin{figure}[htbp]
    \centering
    \includegraphics[width=0.95\linewidth]{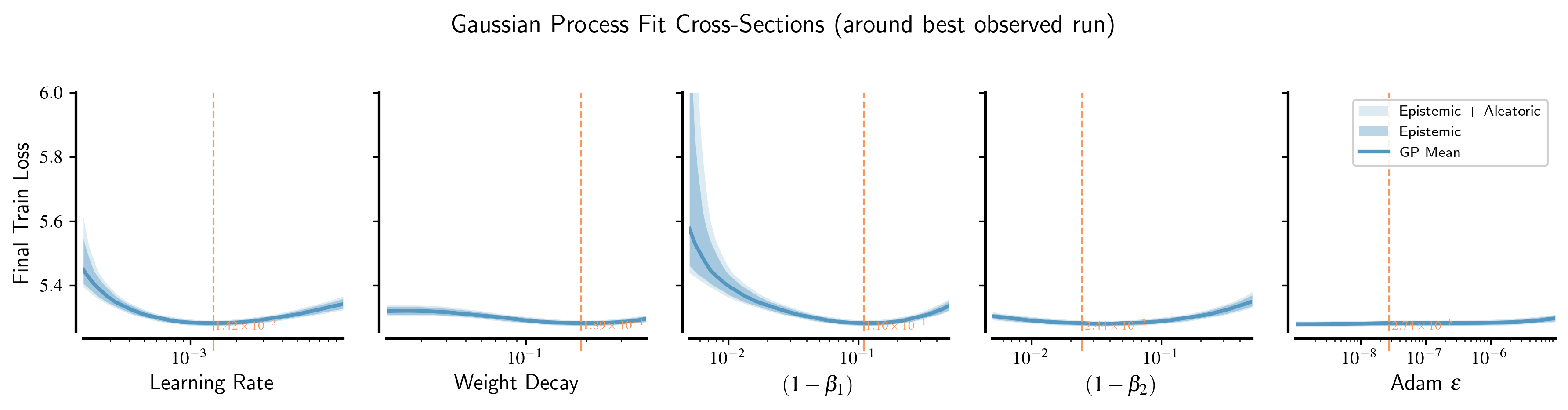}
    \caption{\textbf{Small-scale hyper-parameter search for 0-shot transfer}. Final average cross-entropy loss for a sweep of $5$ AdamW hyperparameters (learning rate, $\beta_1, \beta_2$, weight-decay, $\epsilon$) on a 350M model trained for 13B tokens. The plot shows cross-sections through the hyperparameter landscape through the best found hyperparameters. %
    } 
    \label{fig:adamwoptgaussianprocess}
\end{figure}

As highlighted in \Cref{sec:hyperparameter_transfer}, most ablations in this work rely on optimal global hyperparameters scaled up with CompleteP~\citep{dey2025don}. In \Cref{fig:adamwoptgaussianprocess} we present a Gaussian Process fit
on 2900 trial runs at small scale (320M parameters in total, including 80M non-embedding ones, 13B tokens) to determine optimal global hyperparameters for tri-modal \gls{mdm}. The optimal values are highlighted in red. We initialize the per-module multiplier search from this optimum to better seed the search process.

\subsection{Runtime as Function of Batch Size}
\label{sec:runtime_batch_size}
\begin{figure}[htbp]
    \centering
    \includegraphics[width=0.34\linewidth]{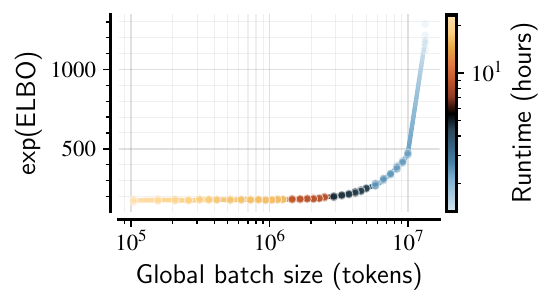}
    \textbf{a)}
    \includegraphics[width=0.29\linewidth]{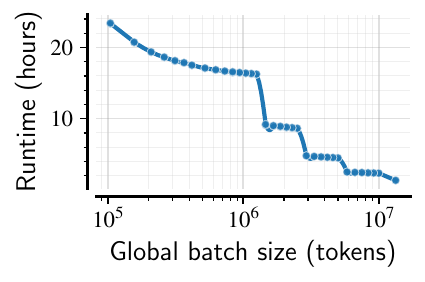}
    \textbf{b)}
    \includegraphics[width=0.29\linewidth]{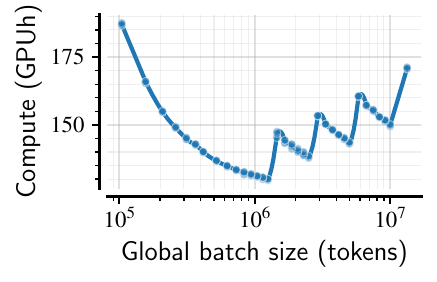}
    \textbf{c)}
    \caption{\textbf{Performance as function of physical batch size.}. In \textbf{a)} we see impact of batch size on total runtime, with the effect of the node-count clearly visible: more nodes reduces wall-clock time faster than simply saturating GPUs. In \textbf{b)} we see the runtime with 1, 2, 4, 8 and 16 nodes respectively. Doubling node count approximately halves the runtime. Finally, in \textbf{c)}, we see that saturated GPUs have much more efficient FLOP usage than non-saturated ones. Doubling node count never allow to recover training efficiency of fewer nodes, because of communication overhead.}
    \label{fig:dpref}
\end{figure}

As explored in~\Cref{sec:scaling}, the SDE parametrization allows a wide range of batch sizes to be used. Typically, bigger batch sizes allow less iterations, which \textit{can} reduce runtime. Two knobs are available to increase the batch size: increasing the number of nodes, or modifying the per-GPU batch size. In practice, overall throughput grows sub-linearly with number of nodes because of communications on the cluster. Furthermore, for smaller batch sizes, hosted on a single node, another phenomenon plays a role: once the batch size is too small, the GPU becomes idle because small batch sizes do not benefit as much from parallelism. This is measured in~\autoref{fig:dpref}. The runtime diminishes slowly as the per-GPU batch size increases, and diminishes sharply as the number of nodes increases. Highest node count reduces wall-clock time significantly, but also reduces FLOP-efficiency because of sub-linear scaling.

\subsection{Hyperparameters for the Unified 3B Tri-modal MDM}
\label{sec:big_model_hps}
\autoref{tab:training_details} highlights all training details for the 3B model.

\begin{table}[htbp]
\centering
\small
\renewcommand{\arraystretch}{0.97}
\begin{tabular}{lc}
\toprule
\multicolumn{2}{c}{\textbf{Model}} \\
\midrule
$N$ blocks & $24$ \\
Dimension & $3072$ \\
$N$ attention heads & $24$ \\
QK Norm & Yes \\
Normalization & RMSNorm \\
Pre-norm & Yes \\
Post-norm & No \\
MLP style & SwiGLU \\
SwiGLU hidden dimension factor & $2.75$ \\
Positional embedding & RoPE \\
Weight initialization (base\_width)& trunc\_normal(std=0.02) \\ %
\midrule
\multicolumn{2}{c}{\textbf{Training parameters}} \\
\midrule
Batch size & $3072$ \\
Sequence length & $3256$ \\
Optimizer & AdamW \\
Base LR & $9e-4$ \\
Base AdamW $\epsilon$ & $1e-8$ \\
Base AdamW $\beta_1$ & $0.9$ \\
Base AdamW $\beta_2$ & $0.95$ \\
Base weight decay & $0.1$ \\
LR warmup & $2,000$ steps \\
LR schedule & Cosine \\
Min LR & $1e-6$ \\
Z-loss weight & $1e-5$ \\
Training duration & $1,000,000$ steps \\
Hyperparameter Transfer Strategy & CompleteP~\citep{dey2025don}] \\
Modality sampling rate (text-only, image-text, audio-text) & $[0.33, 0.33, 0.33]$ \\
Text tokens seen during training & 3.4T \\
Image samples seen during training & 1B \\
Audio samples seen during training & 1B\\
\midrule
\multicolumn{2}{c}{\textbf{Tokenizers}} \\
\midrule
Text & Tiktoken \citep{openai_tiktoken} \\
Image & SBER-MoVQGAN \citep{sber_movqgan} \\
Audio & Higgs Audio Tokenizer v2 (4 codebooks) \citep{higgsaudio2025} \\
\midrule
\multicolumn{2}{c}{\textbf{Vocabulary (incl. special tokens)}} \\
\midrule
Total & $117,698$ \\
Text & $100,281$ \\
Image & $16,387$ \\
Audio & $1,027$ \\
\midrule
\multicolumn{2}{c}{\textbf{Text transformations}} \\
\midrule
$P(\text{Token packing})$ & 0.95 \\
$P(\text{Random sequence subsample})$ & 0.05 \\
\midrule
\multicolumn{2}{c}{\textbf{Image transformations}} \\
\midrule
Target resolution & $256x256$ \\
RandomResizedCrop & $[0.8, 1.0]$ \\
Resize with white padding & - \\
$P(\text{RandomResizedCrop})$ & $0.5$ \\
$P(\text{Resize with white padding})$ & $0.5$ \\
Normalization $\mu$ and $\sigma$ & $[0.5, 0.5]$ \\
\midrule
\multicolumn{2}{c}{\textbf{Audio transformations}} \\
\midrule
Max duration & $30$ seconds \\
Number of frames & $25$ \\
\bottomrule
\end{tabular}
\caption{Model and training details for the 3B multimodal \gls{mdm}.}
\label{tab:training_details}
\end{table}

\section{MDM with Per-module Hyperparameters}
\label{app:per_module_hyperparam}
To simplify and parallelize our analysis, we rely on CompleteP~\citep{dey2025don} for width and depth transfer using global hyperparameters (Appendix \Cref{sec:opt_globals}) for all ablations done in this work, however, recent insights from \cite{mlodozeniec2025completed} highlight that we can further improve performance by optimizing per-module hyper-parameter multipliers for AdamW (learning rate, weight decay, $\beta_1, \beta_2$, and $\epsilon$).

\begin{figure}[htbp]
    \centering
    \includegraphics[width=0.75\linewidth]{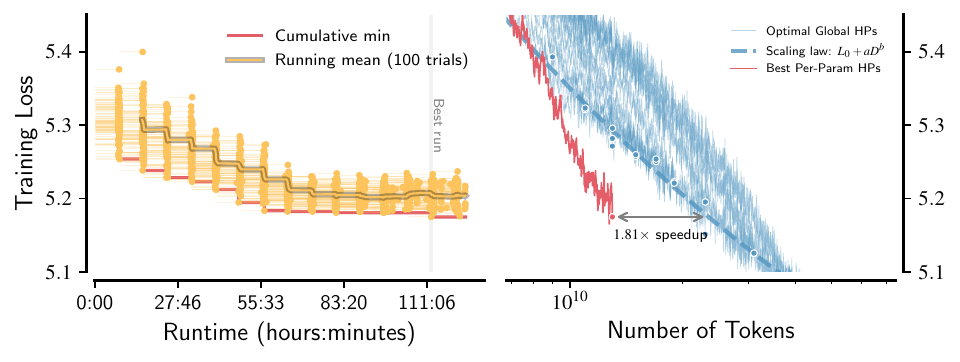}
    \caption{\textbf{Per-module hyperparameter search on a small scale model for 0-shot transfer}. Average final ELBO for a sweep of AdamW's per-module hyperparameters as well as initialization scales for a 350M model (including 80M non-embedding parameters) trained for 13B tokens, using a batch size of 256. Per-module tuning yields a \textbf{1.81}$\times$ reduction in token count to achieve an equivalent loss.
    }
    \label{fig:per-module-search}
\end{figure}

\paragraph{Training Details.} Each hyperparameter for each weight gets a unique multiplier. The multipliers for all depth-repeated blocks are parameterized as a product of a depth-dependent factor, and a module-type factor – all weights at the same depth get the same depth factor, and all weights of the same type (e.g., all $QKV$ weights) get the same module-type factor. We initialize the local random search method from \citep{mlodozeniec2025completed} at the best global hyperparameters found using random search, shown in ~\autoref{fig:adamwoptgaussianprocess}. We conduct the search using a transformer with 8 blocks of width 1024, totaling 320M parameters (including 80M non-embed parameters) trained at a horizon of 13B multimodal tokens. The results of the search, and the speed-up obtained at this base model size, is reported in \autoref{fig:per-module-search}. 

\begin{table}[htbp]
\centering
\caption{Optimal per-module hyperparameter multipliers found for the LLaDA Multimodal model. Depth factors apply to all layers where the $\frac{\texttt{block depth}}{\texttt{total depth}}$ falls within the highlighted fraction (as counted from the network input towards output).}\label{tab:per-module-multipliers}
\vspace{0.5cm}
\label{tab:hp_multipliers}
\begin{tabular}{llcccccc}
\toprule
\textbf{Category} & \textbf{Module / Depth} & \textbf{LR} & \textbf{WD} & \textbf{$\alpha_1$} & \textbf{$\alpha_2$} & \textbf{$\epsilon$} & \textbf{Init Scale} \\
\midrule
\multirow{7}{*}{Standalone}
& Embedding weights (Audio) & 2.192 & 1.009 & 0.962 & 1.493 & 1.494 & 1.826 \\
& Embedding weights (Image) & 1.013 & 0.864 & 2.108 & 0.685 & 0.734 & 0.554 \\
& Embedding weight (Text) & 3.937 & 1.593 & 1.421 & 1.791 & 0.317 & 0.379 \\
\cmidrule{2-8}
& Unembedding weights (Audio) & 1.633 & 1.510 & 3.442 & 0.594 & 0.742 & 3.422 \\
& Unembedding weights (Image) & 1.655 & 1.213 & 1.929 & 1.042 & 0.635 & 1.524 \\
& Unembedding weights (Text) & 3.008 & 0.737 & 1.346 & 0.955 & 1.206 & 0.341 \\
& Unembedding norm weights & 2.305 & 0.817 & 4.508 & 2.740 & 1.938 & 2.175 \\
\midrule
\multirow{4}{*}{Blocks}
& \texttt{attn\_qkv\_weight} & 1.714 & 0.821 & 0.173 & 0.557 & 0.391 & 2.498 \\
& \texttt{attn\_proj\_weight} & 0.630 & 0.354 & 0.256 & 0.339 & 1.627 & 4.732 \\
& \texttt{attn\_q\_norm\_weight} & 0.535 & 0.731 & 1.530 & 0.902 & 0.848 & 1.344 \\
& \texttt{attn\_k\_norm\_weight} & 0.754 & 0.497 & 1.074 & 0.822 & 0.368 & 0.436 \\
\cmidrule{2-8}
& \texttt{mlp\_gate\_weight} & 0.489 & 0.634 & 1.171 & 1.870 & 4.913 & 0.643 \\
& \texttt{mlp\_fc1\_weight} & 1.271 & 1.295 & 1.590 & 3.309 & 1.415 & 1.944 \\
& \texttt{mlp\_fc2\_weight} & 1.405 & 1.308 & 2.684 & 0.655 & 1.790 & 0.878 \\
\cmidrule{2-8}
& \texttt{norm1\_weight} & 1.311 & 1.105 & 0.282 & 1.161 & 1.477 & 2.171 \\
& \texttt{norm2\_weight} & 0.899 & 0.525 & 1.533 & 1.789 & 0.712 & 1.189 \\
\midrule
\multirow{2}{*}{Depth Factors}
& $0-50\%$ & 1.102 & 0.725 & 1.030 & 3.053 & 0.663 & 0.997 \\
& $50-100\%$ & 0.877 & 1.018 & 0.911 & 1.149 & 2.645 & 0.485 \\
\bottomrule
\end{tabular}
\end{table}

\paragraph{Multiplier analysis.} In~\Cref{tab:hp_multipliers}, we highlight the results of the search, listing the per module multipliers. Notably, the resulting multipliers are highly structured rather than uniform: embedding and unembedding weights favor substantially larger effective learning rates (up to $\sim\!4\times$), while attention projections and MLP gates are tuned more conservatively, often with increased $\epsilon$ for numerical damping. The learned depth factors further indicate smaller steps and stronger stabilization in later blocks, consistent with increasing sensitivity of deep representations to update noise.

\FloatBarrier

\section{Extended Scaling Laws Results}
\label{sec:extended_scaling_laws}

\paragraph{Computation of FLOPs for experiments.} We rely on the formula in appendix \textbf{H.1} of~\citet{DBLP:conf/icml/BusbridgeSWRLW25} to compute the FLOPs and the model size. In particular, we do not account for input/output embedding, as the size of our multimodal vocabulary (117k) makes these embedding matrices bigger than the transformer backbone for small models. The ratio between transformer parameters and embedding parameters is illustrated in~\autoref{fig:ratiorealfakeparams}. The value of $N$ reported in scaling laws fit always use the non-embedding parameters only (which are up-to 5 times smaller than the total model's size). The FLOPs $C$ reported in scaling law account for everything. We compute the optimal $N^{\star}(C)$ and $D^{\star}(C)$ by minimizing the parametric loss under the constraint $C=\text{FLOP per token}(N)\times D$. Since we do not account for embeddings parameters in total model size, the popular $\text{FLOP per token}(N)=6N$ does not apply out-of-the-box. Instead, we minimize the parametric loss $L(N,D=C/\text{FLOP per token}(N))$ via linear-search over $N$ to plot compute-optimal curves in~\autoref{fig:pareto-front-flops} and~\autoref{fig:trimodalisoflops}. This effect is more striking for small models where the embedding size is significant. For larger models, we found that the approximation $\text{FLOP per token}(N)\approx 6$ holds. In that case, minimizing the parametric form
\begin{equation*}
    L(N, D)=E+\left(AN^{-a/b}+BD^{-1}\right)^{b},
\end{equation*}
under constraint $C=6ND$ works well. By monotonicity, this is equivalent to minimizing $AN^{-a/b}+BD^{-1}$, which admits the following minimizers:
\begin{equation*}
    N^{\star}(C)= G^{-1}(C/6)^{\tau},\quad\text{ and }\quad D^{\star}(C)= G(C/6)^{1-\tau},\quad \text{ with }\quad G=\frac{bB}{aA}\quad\text{ and }\quad\tau=\frac{b}{a+b}.  
\end{equation*}

\begin{figure}[ht]
    \centering
    \includegraphics[width=1\linewidth]{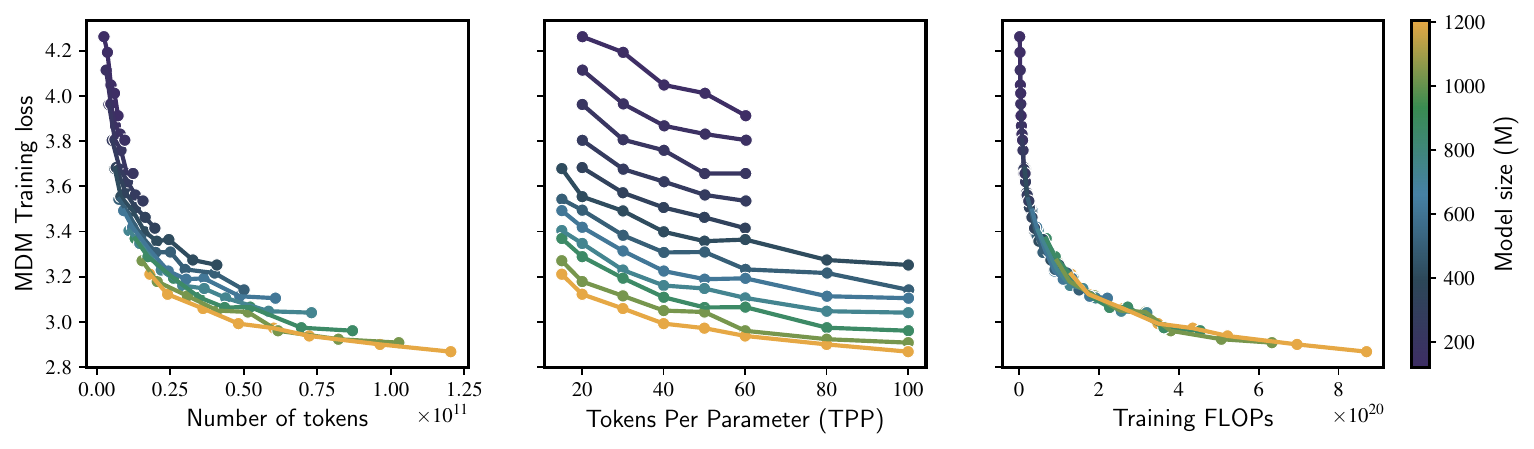}
    \caption{\textbf{Training curves for uni-modal text \gls{mdm} models trained under CompleteP.}}%
    \label{fig:trainingtextscaling}
    \vspace{-5pt}
\end{figure}

\begin{figure}[!ht]
    \centering
    \textbf{a)}
    \includegraphics[width=0.38\linewidth]{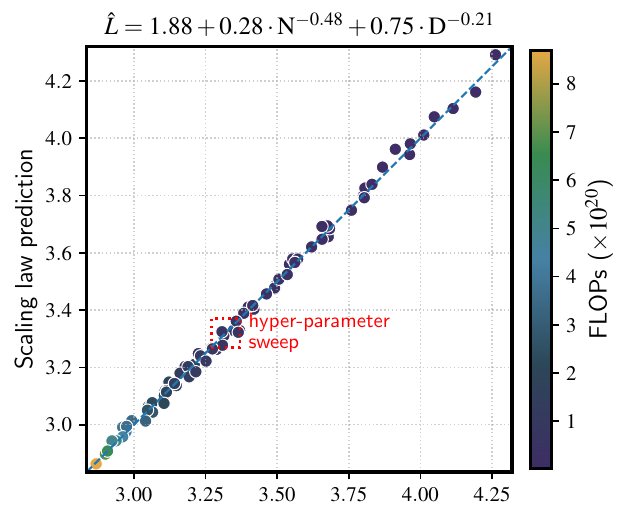}
    \textbf{b)}
    \includegraphics[width=0.45\linewidth]{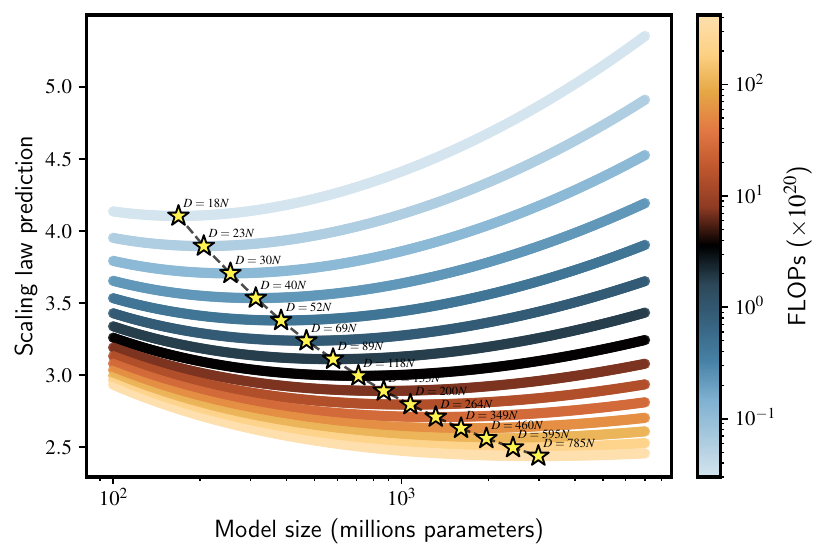}
    \caption{\textbf{Scaling laws for uni-modal text \gls{mdm}s}. \textbf{a)} Scaling law predictions for text-only \gls{mdm} models, using CompleteP. \textbf{b)} Iso-FLOP curves for text-only \gls{mdm} models under CompleteP parameterization (no SDE scaling).}  
    \label{fig:textscaling}
\end{figure}

\paragraph{Scaling laws for Uni-Modal Text MDM.} We also perform scaling laws run on uni-modal text models, using CompleteP parametrization, but \textit{without} SDE scaling rules. Every model size relies on a different batch size to maximize GPU occupancy. The total sequence length is 4,096 with packing and truncation (no padding). Training curves as function of $(N,D)$ FLOP budget are given in~\autoref{fig:trainingtextscaling} and the scaling laws predictions are reported in~\autoref{fig:textscaling}. 

\section{Masking Schedules for Image and Audio Generation}
\label{sec:masking_schedule_ablation}
\begin{figure}[ht]
\centering
\scalebox{0.8}{
\includegraphics[width=\linewidth]{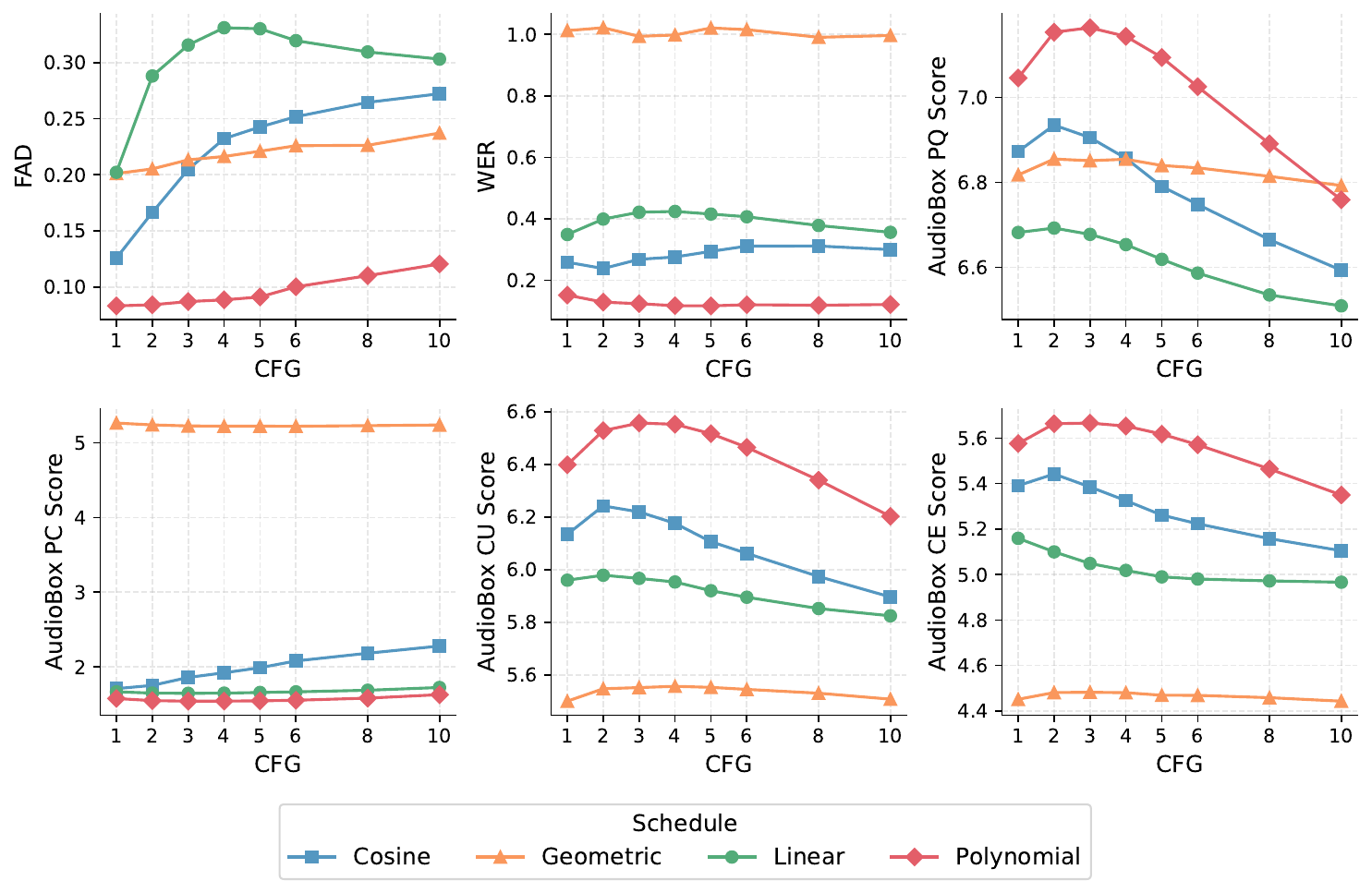}
}
\caption{\textbf{Masking schedule ablation for audio generation across guidance scales.} We evaluate four masking schedules (linear, cosine, polynomial, geometric) on ground-truth length audio generation quality using FAD, WER, and AudioBox Aesthetics metrics on our train mixture. Models are evaluated across CFG scales 1.0-10.0.}
\label{fig:audio_schedule_ablation}
\end{figure}
\begin{figure}[ht]
\centering
\includegraphics[width=\linewidth]{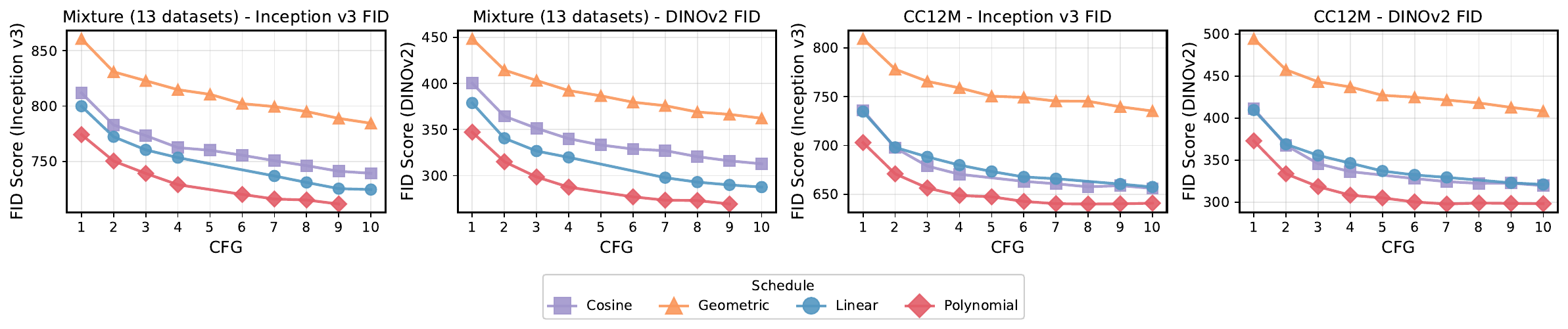}
\caption{\textbf{Masking schedule ablation across guidance scales.} We evaluate the generation quality of four masking schedules (linear, cosine, polynomial, geometric) in CC12M and our train mixture (eval seed). Models are evaluated across CFG scales 1.0-10.0.}
\label{fig:schedule_ablation}
\end{figure}
To determine the impact of the masking schedule on multimodal \gls{mdm} training and generation quality, we evaluate four distinct schedules -- linear, cosine, polynomial, and geometric -- implemented using the continuous-time ELBO weighting in ~\citet{DBLP:conf/nips/ShiHWDT24}. We train a 1B model for 100k steps under each schedule, keeping all other hyperparameters the same.

First, we evaluate image generation quality at $256\times256$ resolution using 1024 diffusion steps with CFG scales ranging from 1.0 to 10.0, temperature $T=0.9$, and nucleus sampling (top-$p=0.9$). Image quality is measured using both FID-Inception and FID-DINOv2, computed over 8,192 generated samples on two datasets, CC12M and our train mixture (eval seed). \autoref{fig:schedule_ablation} shows that the polynomial schedule consistently achieves the best image quality across both metrics and datasets among the four schedules tested. Both metrics agree that polynomial yields superior generation quality, with optimal performance in the CFG range of 7 to 9.

Then, we evaluate audio generation quality with ground-truth durations using 512 diffusion steps with CFG scales ranging from 1.0 to 10.0, temperature $T=1.0$, and nucleus sampling (top-$p=0.9$). We measure audio quality using FAD, WER, and AudioBox Aesthetics computed over 10,000 generated samples from the dataset. \autoref{fig:audio_schedule_ablation} shows that the polynomial schedule also consistently achieves the best audio generation quality across all six metrics evaluated. Unlike in image generation, the optimal CFG range is between 1 and 3.

\newcounter{globalpanel}
\setcounter{globalpanel}{0} %
\newcommand{\panelcaption}{%
  \refstepcounter{globalpanel}%
  \caption*{\textit{(\alphalph{\value{globalpanel}})}}%
}

\section{Extended generations}
\label{app:extended_generations}

\begin{figure*}[t]
    \centering

    \begin{subfigure}{0.22\textwidth}\includegraphics[width=\linewidth]{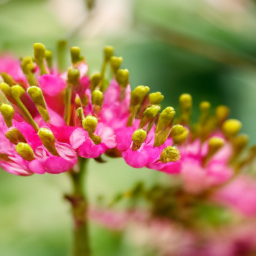}\panelcaption\end{subfigure}\hfill
    \begin{subfigure}{0.22\textwidth}\includegraphics[width=\linewidth]{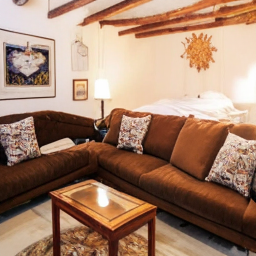}\panelcaption\end{subfigure}\hfill
    \begin{subfigure}{0.22\textwidth}\includegraphics[width=\linewidth]{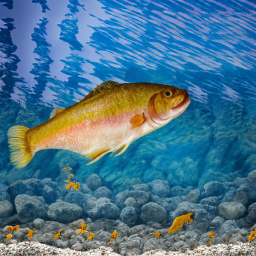}\panelcaption\end{subfigure}\hfill
    \begin{subfigure}{0.22\textwidth}\includegraphics[width=\linewidth]{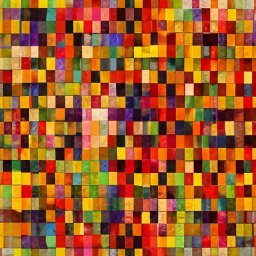}\panelcaption\end{subfigure}

    \vspace{0.5em}

    \begin{subfigure}{0.22\textwidth}\includegraphics[width=\linewidth]{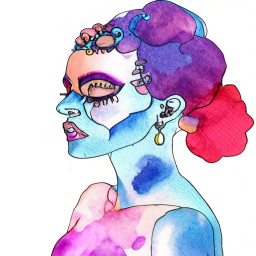}\panelcaption\end{subfigure}\hfill
    \begin{subfigure}{0.22\textwidth}\includegraphics[width=\linewidth]{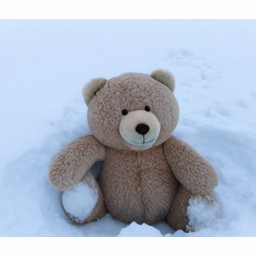}\panelcaption\end{subfigure}\hfill
    \begin{subfigure}{0.22\textwidth}\includegraphics[width=\linewidth]{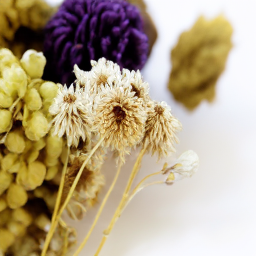}\panelcaption\end{subfigure}\hfill
    \begin{subfigure}{0.22\textwidth}\includegraphics[width=\linewidth]{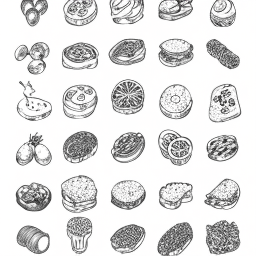}\panelcaption\end{subfigure}

    \vspace{0.5em}

    \begin{subfigure}{0.22\textwidth}\includegraphics[width=\linewidth]{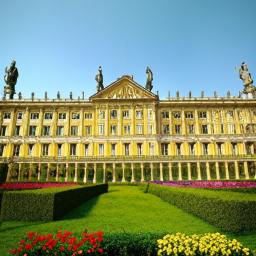}\panelcaption\end{subfigure}\hfill
    \begin{subfigure}{0.22\textwidth}\includegraphics[width=\linewidth]{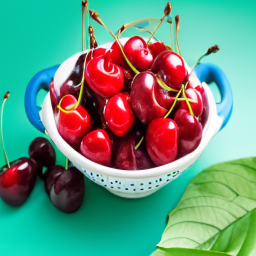}\panelcaption\end{subfigure}
    \begin{subfigure}{0.22\textwidth}\includegraphics[width=\linewidth]{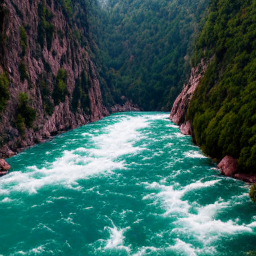}\panelcaption\end{subfigure}\hfill
    \begin{subfigure}{0.22\textwidth}\includegraphics[width=\linewidth]{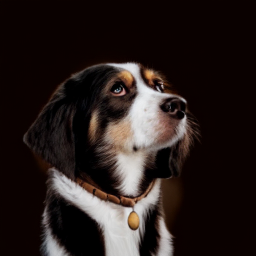}\panelcaption\end{subfigure}\hfill

    \vspace{0.5em}

    \begin{subfigure}{0.22\textwidth}\includegraphics[width=\linewidth]{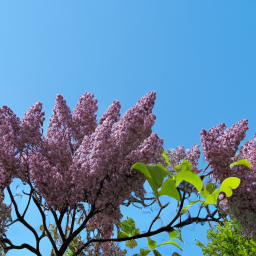}\panelcaption\end{subfigure}\hfill
    \begin{subfigure}{0.22\textwidth}\includegraphics[width=\linewidth]{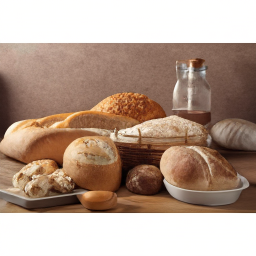}\panelcaption\end{subfigure}\hfill
    \begin{subfigure}{0.22\textwidth}\includegraphics[width=\linewidth]{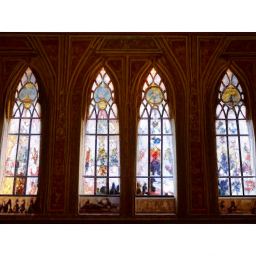}\panelcaption\end{subfigure}\hfill
    \begin{subfigure}{0.22\textwidth}\includegraphics[width=\linewidth]{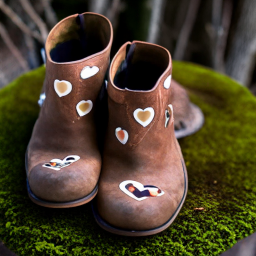}\panelcaption\end{subfigure}

    \vspace{0.5em}

    \begin{subfigure}{0.22\textwidth}\includegraphics[width=\linewidth]{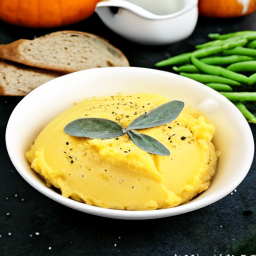}\panelcaption\end{subfigure}\hfill
    \begin{subfigure}{0.22\textwidth}\includegraphics[width=\linewidth]{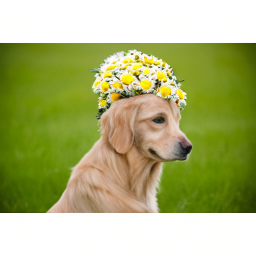}\panelcaption\end{subfigure}\hfill
    \begin{subfigure}{0.22\textwidth}\includegraphics[width=\linewidth]{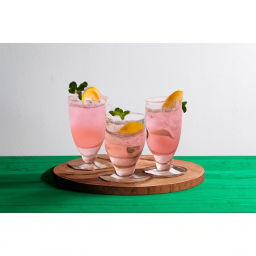}\panelcaption\end{subfigure}\hfill
    \begin{subfigure}{0.22\textwidth}\includegraphics[width=\linewidth]{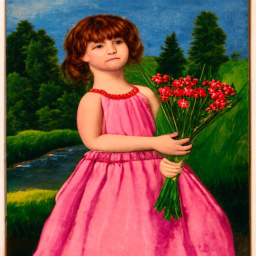}\panelcaption\end{subfigure}

    \caption{
        \textbf{Samples generated by our model with different prompts. See \autoref{tab:generated_images_prompts_1} for the extensive list of prompts.}
    }
    \label{fig:extended_image_generations_1}
\end{figure*}

\begin{figure*}[t]
    \centering

    \begin{subfigure}{0.22\textwidth}\includegraphics[width=\linewidth]{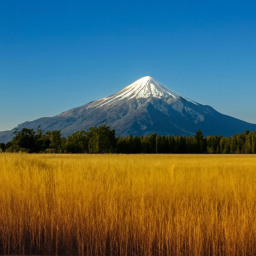}\panelcaption\end{subfigure}\hfill
    \begin{subfigure}{0.22\textwidth}\includegraphics[width=\linewidth]{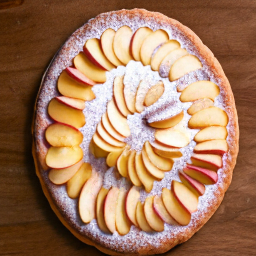}\panelcaption\end{subfigure}\hfill
    \begin{subfigure}{0.22\textwidth}\includegraphics[width=\linewidth]{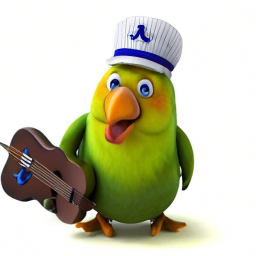}\panelcaption\end{subfigure}\hfill
    \begin{subfigure}{0.22\textwidth}\includegraphics[width=\linewidth]{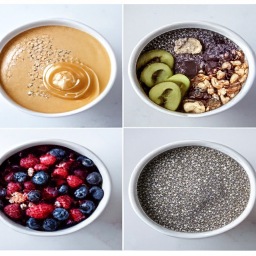}\panelcaption\end{subfigure}

    \vspace{0.5em}

    \begin{subfigure}{0.22\textwidth}\includegraphics[width=\linewidth]{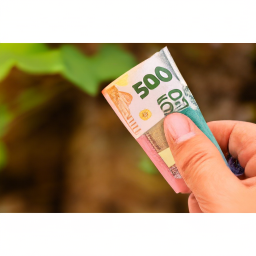}\panelcaption\end{subfigure}\hfill
    \begin{subfigure}{0.22\textwidth}\includegraphics[width=\linewidth]{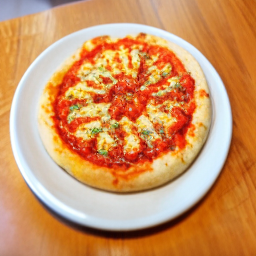}\panelcaption\end{subfigure}\hfill
    \begin{subfigure}{0.22\textwidth}\includegraphics[width=\linewidth]{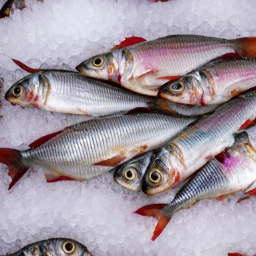}\panelcaption\end{subfigure}\hfill
    \begin{subfigure}{0.22\textwidth}\includegraphics[width=\linewidth]{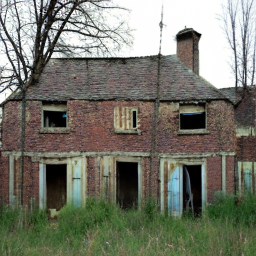}\panelcaption\end{subfigure}

    \vspace{0.5em}

    \begin{subfigure}{0.22\textwidth}\includegraphics[width=\linewidth]{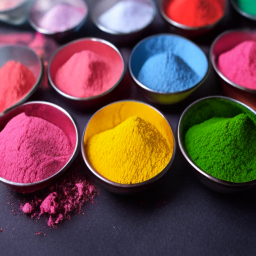}\panelcaption\end{subfigure}\hfill
    \begin{subfigure}{0.22\textwidth}\includegraphics[width=\linewidth]{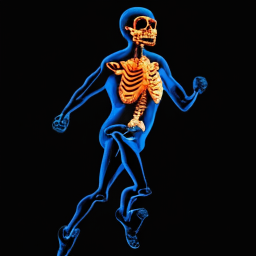}\panelcaption\end{subfigure}
    \begin{subfigure}{0.22\textwidth}\includegraphics[width=\linewidth]{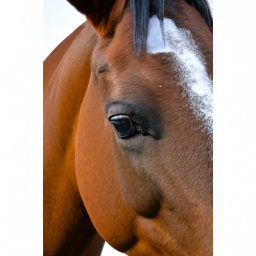}\panelcaption\end{subfigure}\hfill
    \begin{subfigure}{0.22\textwidth}\includegraphics[width=\linewidth]{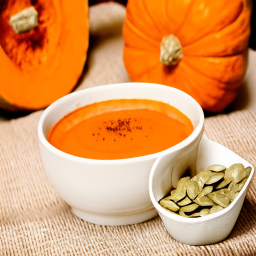}\panelcaption\end{subfigure}\hfill

    \vspace{0.5em}

    \begin{subfigure}{0.22\textwidth}\includegraphics[width=\linewidth]{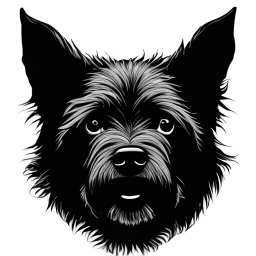}\panelcaption\end{subfigure}\hfill
    \begin{subfigure}{0.22\textwidth}\includegraphics[width=\linewidth]{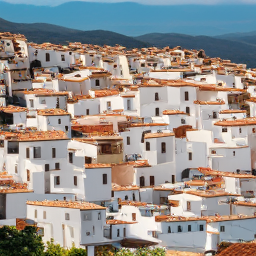}\panelcaption\end{subfigure}\hfill
    \begin{subfigure}{0.22\textwidth}\includegraphics[width=\linewidth]{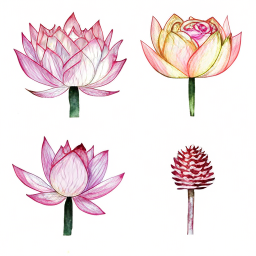}\panelcaption\end{subfigure}\hfill
    \begin{subfigure}{0.22\textwidth}\includegraphics[width=\linewidth]{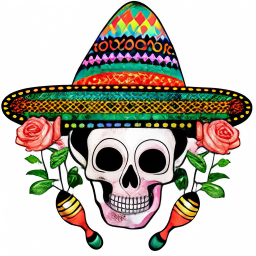}\panelcaption\end{subfigure}

    \vspace{0.5em}

    \begin{subfigure}{0.22\textwidth}\includegraphics[width=\linewidth]{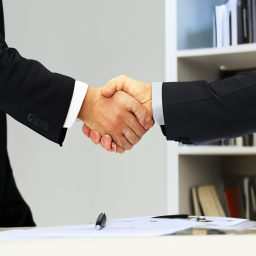}\panelcaption\end{subfigure}\hfill
    \begin{subfigure}{0.22\textwidth}\includegraphics[width=\linewidth]{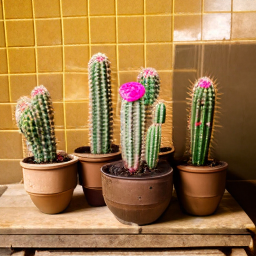}\panelcaption\end{subfigure}\hfill
    \begin{subfigure}{0.22\textwidth}\includegraphics[width=\linewidth]{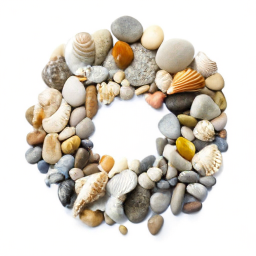}\panelcaption\end{subfigure}\hfill
    \begin{subfigure}{0.22\textwidth}\includegraphics[width=\linewidth]{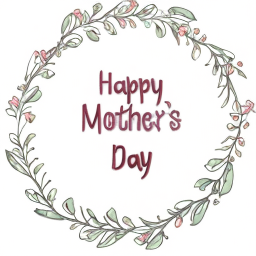}\panelcaption\end{subfigure}

    \caption{
        \textbf{Samples generated by our model with different prompts.}
    }
    \label{fig:extended_image_generations_2}
\end{figure*}

\begin{figure*}[t]
    \centering

    \begin{subfigure}{0.22\textwidth}\includegraphics[width=\linewidth]{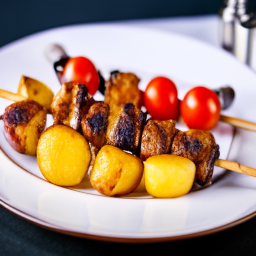}\panelcaption\end{subfigure}\hfill
    \begin{subfigure}{0.22\textwidth}\includegraphics[width=\linewidth]{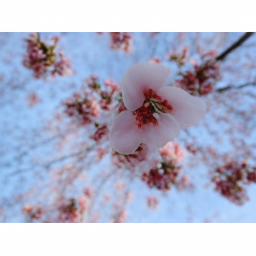}\panelcaption\end{subfigure}\hfill
    \begin{subfigure}{0.22\textwidth}\includegraphics[width=\linewidth]{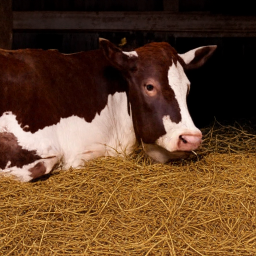}\panelcaption\end{subfigure}\hfill
    \begin{subfigure}{0.22\textwidth}\includegraphics[width=\linewidth]{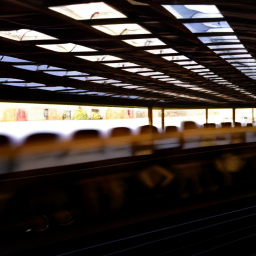}\panelcaption\end{subfigure}

    \vspace{0.5em}

    \begin{subfigure}{0.22\textwidth}\includegraphics[width=\linewidth]{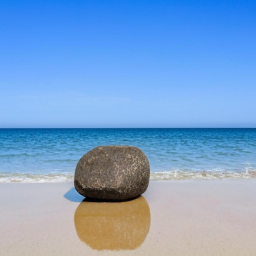}\panelcaption\end{subfigure}\hfill
    \begin{subfigure}{0.22\textwidth}\includegraphics[width=\linewidth]{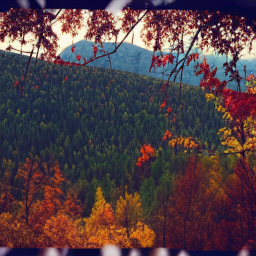}\panelcaption\end{subfigure}\hfill
    \begin{subfigure}{0.22\textwidth}\includegraphics[width=\linewidth]{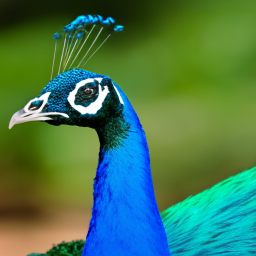}\panelcaption\end{subfigure}\hfill
    \begin{subfigure}{0.22\textwidth}\includegraphics[width=\linewidth]{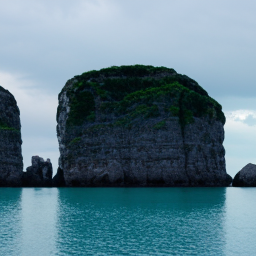}\panelcaption\end{subfigure}

    \vspace{0.5em}

    \begin{subfigure}{0.22\textwidth}\includegraphics[width=\linewidth]{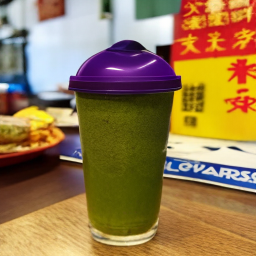}\panelcaption\end{subfigure}\hfill
    \begin{subfigure}{0.22\textwidth}\includegraphics[width=\linewidth]{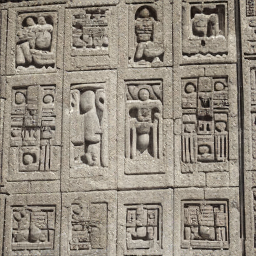}\panelcaption\end{subfigure}
    \begin{subfigure}{0.22\textwidth}\includegraphics[width=\linewidth]{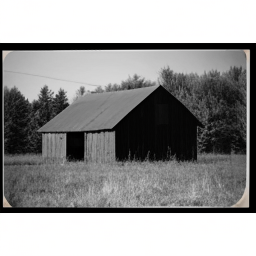}\panelcaption\end{subfigure}\hfill
    \begin{subfigure}{0.22\textwidth}\includegraphics[width=\linewidth]{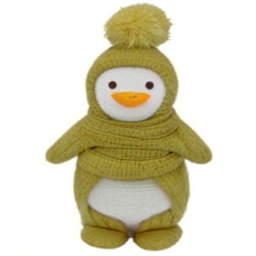}\panelcaption\end{subfigure}\hfill

    \vspace{0.5em}

    \begin{subfigure}{0.22\textwidth}\includegraphics[width=\linewidth]{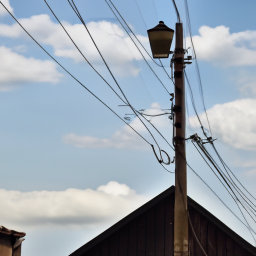}\panelcaption\end{subfigure}\hfill
    \begin{subfigure}{0.22\textwidth}\includegraphics[width=\linewidth]{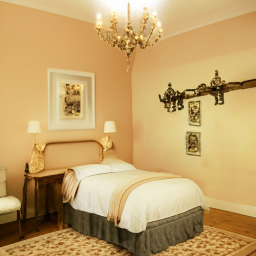}\panelcaption\end{subfigure}\hfill
    \begin{subfigure}{0.22\textwidth}\includegraphics[width=\linewidth]{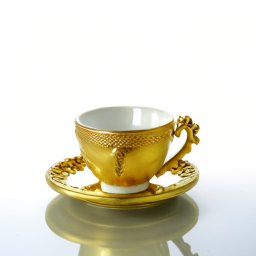}\panelcaption\end{subfigure}\hfill
    \begin{subfigure}{0.22\textwidth}\includegraphics[width=\linewidth]{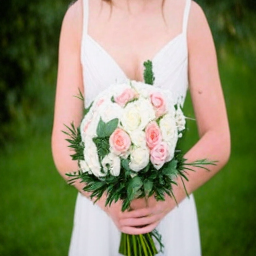}\panelcaption\end{subfigure}

    \vspace{0.5em}

    \begin{subfigure}{0.22\textwidth}\includegraphics[width=\linewidth]{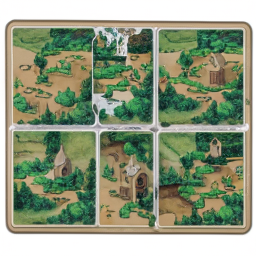}\panelcaption\end{subfigure}\hfill
    \begin{subfigure}{0.22\textwidth}\includegraphics[width=\linewidth]{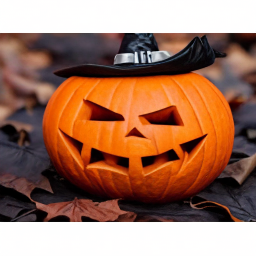}\panelcaption\end{subfigure}\hfill
    \begin{subfigure}{0.22\textwidth}\includegraphics[width=\linewidth]{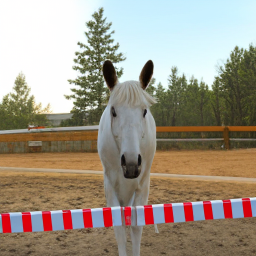}\panelcaption\end{subfigure}\hfill
    \begin{subfigure}{0.22\textwidth}\includegraphics[width=\linewidth]{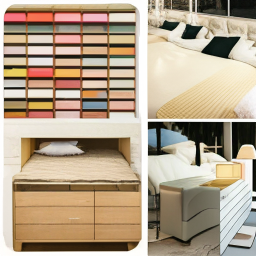}\panelcaption\end{subfigure}

    \caption{
        \textbf{Samples generated by our model with different prompts.}
    }
    \label{fig:extended_image_generations_3}
\end{figure*}

\begin{figure*}[t]
    \centering

    \begin{subfigure}{0.22\textwidth}\includegraphics[width=\linewidth]{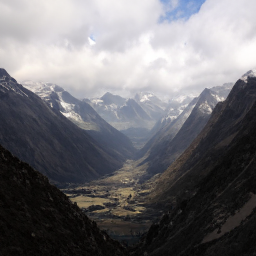}\panelcaption\end{subfigure}\hfill
    \begin{subfigure}{0.22\textwidth}\includegraphics[width=\linewidth]{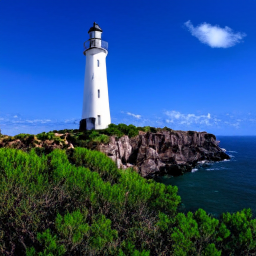}\panelcaption\end{subfigure}\hfill
    \begin{subfigure}{0.22\textwidth}\includegraphics[width=\linewidth]{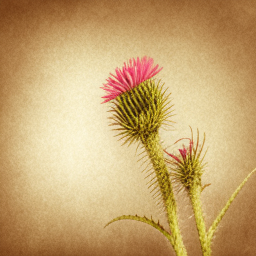}\panelcaption\end{subfigure}\hfill
    \begin{subfigure}{0.22\textwidth}\includegraphics[width=\linewidth]{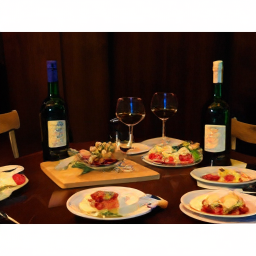}\panelcaption\end{subfigure}

    \vspace{0.5em}

    \begin{subfigure}{0.22\textwidth}\includegraphics[width=\linewidth]{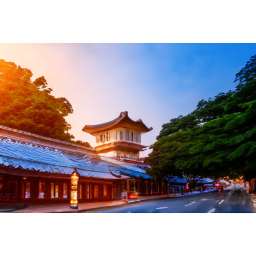}\panelcaption\end{subfigure}\hfill
    \begin{subfigure}{0.22\textwidth}\includegraphics[width=\linewidth]{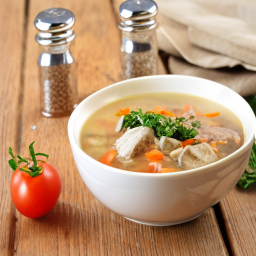}\panelcaption\end{subfigure}\hfill
    \begin{subfigure}{0.22\textwidth}\includegraphics[width=\linewidth]{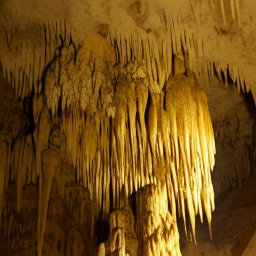}\panelcaption\end{subfigure}\hfill
    \begin{subfigure}{0.22\textwidth}\includegraphics[width=\linewidth]{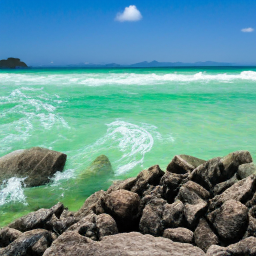}\panelcaption\end{subfigure}

    \vspace{0.5em}

    \begin{subfigure}{0.22\textwidth}\includegraphics[width=\linewidth]{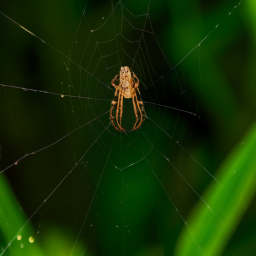}\panelcaption\end{subfigure}\hfill
    \begin{subfigure}{0.22\textwidth}\includegraphics[width=\linewidth]{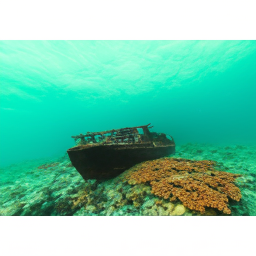}\panelcaption\end{subfigure}
    \begin{subfigure}{0.22\textwidth}\includegraphics[width=\linewidth]{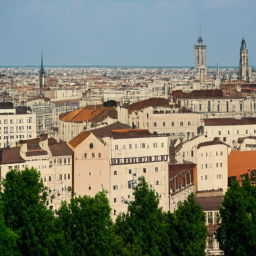}\panelcaption\end{subfigure}\hfill
    \begin{subfigure}{0.22\textwidth}\includegraphics[width=\linewidth]{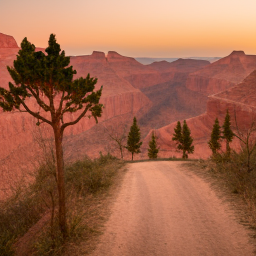}\panelcaption\end{subfigure}\hfill

    \vspace{0.5em}

    \begin{subfigure}{0.22\textwidth}\includegraphics[width=\linewidth]{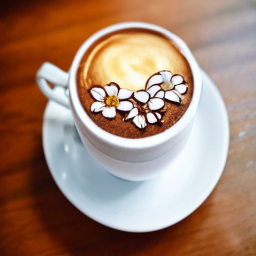}\panelcaption\end{subfigure}\hfill
    \begin{subfigure}{0.22\textwidth}\includegraphics[width=\linewidth]{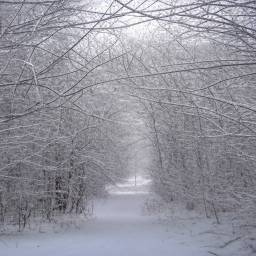}\panelcaption\end{subfigure}\hfill
    \begin{subfigure}{0.22\textwidth}\includegraphics[width=\linewidth]{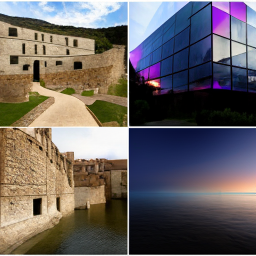}\panelcaption\end{subfigure}\hfill
    \begin{subfigure}{0.22\textwidth}\includegraphics[width=\linewidth]{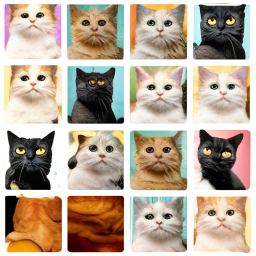}\panelcaption\end{subfigure}

    \vspace{0.5em}

    \begin{subfigure}{0.22\textwidth}\includegraphics[width=\linewidth]{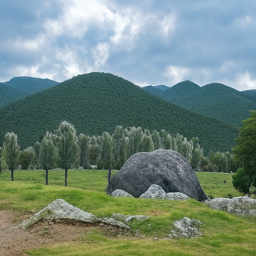}\panelcaption\end{subfigure}\hfill
    \begin{subfigure}{0.22\textwidth}\includegraphics[width=\linewidth]{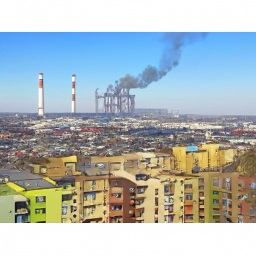}\panelcaption\end{subfigure}\hfill
    \begin{subfigure}{0.22\textwidth}\includegraphics[width=\linewidth]{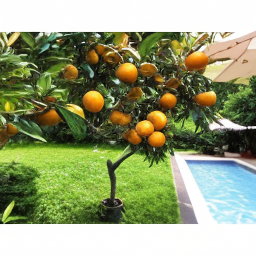}\panelcaption\end{subfigure}\hfill
    \begin{subfigure}{0.22\textwidth}\includegraphics[width=\linewidth]{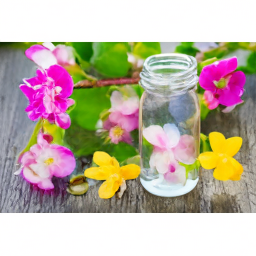}\panelcaption\end{subfigure}

    \caption{
        \textbf{Samples generated by our model with different prompts.}
    }
    \label{fig:extended_image_generations_4}
\end{figure*}

We present extended generations in \autoref{fig:extended_image_generations_1}, \autoref{fig:extended_image_generations_2}, \autoref{fig:extended_image_generations_3}, and \autoref{fig:extended_image_generations_4} and their respective complete list of prompts in \autoref{tab:generated_images_prompts_1}, \autoref{tab:generated_images_prompts_2}, \autoref{tab:generated_images_prompts_3}, and \autoref{tab:generated_images_prompts_4}. Note that the prompts come from synthetic captions and were selected among a larger set of generations based a mix of quality filtering and diversity.

\FloatBarrier

\begin{table*}[t]
    \centering
    \renewcommand{\arraystretch}{1}
    \begin{tabular}{p{0.98\textwidth}}
    \hline
        \textbf{(a)} Lagerstroemia macrocarpa Wall, Queen's flower, Queen's crape myrtle, Lythraceae. Close-up of a vibrant pink flower with a yellow center in full bloom, sharply focused against a blurred green outdoor background. Numerous small green buds at different stages surround the flower on thin green stems. \\ \hline
        \textbf{(b)} A cozy living room with a plush brown sofa and patterned pillows. A wooden coffee table sits in front, framed wall art above, rustic wooden ceiling, and a bed with a white comforter in the background. Decorative lighting and wall hangings add warmth. \\ \hline
        \textbf{(c)} A photograph of a rainbow trout in a body of water. The trout is facing the left side of the image. The water is clear and there are some small rocks at the bottom of the water. There are also some small fish in the water. \\ \hline
        \textbf{(d)} A close-up photograph of a painting with a variety of colors. The colors are bright and vibrant. The painting is made up of small squares of color that are all the same size. \\ \hline
        \textbf{(e)} Watercolor drawing of lady with punk makeup. \\ \hline
        \textbf{(f)} A teddy bear lies alone among the snow. \\ \hline
        \textbf{(g)} A close-up photo of dried white flowers with brown centers, surrounded by purple and green dried flowers. The arrangement is set against a plain white wall, emphasizing texture and muted tones. \\ \hline
        \textbf{(h)} A digital illustration of various food items rendered in black and white, arranged in a grid pattern on a white background. \\ \hline
        \textbf{(i)} A grand palace, likely the Palace of Versailles, with gold doors, statues, columns, and a clock tower. A formal garden with trimmed hedges and colorful flowers stands in front under a clear, sunny sky. \\ \hline
        \textbf{(j)} Fresh red cherries with stems in a white bowl with a blue rim on a teal surface. A few cherries and a green leaf lie outside the bowl, with a softly blurred background emphasizing freshness. \\ \hline
        \textbf{(k)} A wide shot of a fast-flowing blue river with white-water rapids, surrounded by steep rocky cliffs covered in dense green trees. \\ \hline
        \textbf{(l)} A close-up photo of a black and white dog looking up to the right. The dog wears a brown collar with a small bronze pendant against a dark brown background. \\ \hline
        \textbf{(m)} A tree in full bloom with deep lavender lilac flowers and vibrant green leaves, set against a clear blue sky with no clouds. \\ \hline
        \textbf{(n)} Composition with bread and rolls on kitchen table. \\ \hline
        \textbf{(o)} These windows and those in the next photo depict scenes from the Old Testament. \\ \hline
        \textbf{(p)} Brown boots with white heart designs on the toes, resting on a moss-covered ring. A blurred tree trunk and branches appear in the background. \\ \hline
        \textbf{(q)} A white bowl of mashed sweet potatoes garnished with sage, black pepper, and sea salt. In the background are a gravy boat, bread, green beans, and a pumpkin. \\ \hline
        \textbf{(r)} A golden retriever wearing a crown of yellow and white flowers, looking to the right, set in a green grassy field. \\ \hline
        \textbf{(s)} Three pink cocktails on a circular wooden tray, each garnished with lemon slices and mint, set against a white wall and green table. \\ \hline
        \textbf{(t)} A painting of a young girl in a pink dress and red coral necklace holding flowers in a lush landscape with forest and stream, conveying innocence and childhood beauty. \\ \hline
    \end{tabular}
    \caption{Prompts used to generate images in \autoref{fig:extended_image_generations_1}.}
    \label{tab:generated_images_prompts_1}
\end{table*}

\begin{table*}[t]
    \centering
    \renewcommand{\arraystretch}{1}
    \begin{tabular}{p{0.98\textwidth}}
    \hline
        \textbf{(u)} A serene landscape with a snow-capped mountain beneath a clear blue sky. Tall golden grass fills the foreground, distant trees add contrast, and warm lighting suggests early morning or late afternoon. \\ \hline
        \textbf{(v)} A close-up photo of a circular apple pie on a wooden surface, with apples arranged in a circular pattern and dusted with powdered sugar. \\ \hline
        \textbf{(w)} A cartoon parrot wearing a blue-and-white sailor hat with an anchor, smiling while holding a guitar. The parrot has green and yellow feathers, an orange beak, blue eyes, and stands on a white background. \\ \hline
        \textbf{(x)} Five bowls of chia pudding arranged in a grid on a white surface, topped with peanut butter, berries, kiwi with chocolate chips, nuts with honey, and banana with granola under bright lighting. \\ \hline
        \textbf{(y)} A close-up photo of a hand holding a rolled up 20 Euro note. The background is out of focus, but there are green leaves and what appears to be a tree. \\ \hline
        \textbf{(z)} A close-up photo of a small pizza topped with red sauce, cheese, and herbs on a white plate placed on a wooden table. \\ \hline
        \textbf{(aa)} Fresh silver fish with pinkish tones displayed on crushed ice in a supermarket. The fish overlap slightly, with finely shaved white ice keeping them fresh and the background cropped tightly. \\ \hline
        \textbf{(ab)} An abandoned brick building with a sloped roof and chimney. Three doors are boarded up, one is open, and overgrown grass and trees suggest long-term neglect. \\ \hline
        \textbf{(ac)} Metallic bowls arranged in a semi-circle, each filled with vibrant powdered pigments including pink, blue, yellow, and green, photographed in close-up on a dark surface. \\ \hline
        \textbf{(ad)} A blue illustration of a skeleton running. The skeleton is highlighted in orange. The background is black. \\ \hline
        \textbf{(ae)} Expanded horse nostrils with rime on facial hairs \\ \hline
        \textbf{(af)} Rustic pumpkin soup in a white bowl on a woven mat, garnished with black pepper. Pumpkin seeds and a whole pumpkin appear nearby, creating a warm autumnal setting. \\ \hline
        \textbf{(ag)} A black and white illustration of a shaggy dog with pointed ears, wide eyes looking forward, and a slightly open mouth against a white background. \\ \hline
        \textbf{(ah)} A traditional Andalusian white village in Casares, Spain, with tightly packed white buildings and terracotta roofs built along a hillside, surrounded by greenery and lacking modern infrastructure. \\ \hline
        \textbf{(ai)} Watercolor botanical illustrations of a pink lotus and a rose with soft brushstrokes and detailed petals, isolated on a white background in a delicate hand-painted style. \\ \hline
        \textbf{(aj)} A colorful Day of the Dead illustration featuring a skull wearing a sombrero and holding maracas, surrounded by flowers on a clean white background. \\ \hline
        \textbf{(ak)} A handshake between two men in suits, one seated at a desk with papers and a pen, the other standing and smiling, with a bookshelf visible behind them. \\ \hline
        \textbf{(al)} Cactus plants with pink flowers in brown pots placed on a wooden table, set against tiled flooring and a brown-and-yellow tiled wall. \\ \hline
        \textbf{(am)} Stones and shells arranged in the shape of a flower on a plain white background. \\ \hline
        \textbf{(an)} Vintage hand drawn rustic wreath with cute spring flowers and hand written text Happy Mother's Day \\ \hline
    \end{tabular}
    \caption{Prompts used to generate images in \autoref{fig:extended_image_generations_2}.}
    \label{tab:generated_images_prompts_2}
\end{table*}

\begin{table*}[t]
    \centering
    \renewcommand{\arraystretch}{1}
    \begin{tabular}{p{0.98\textwidth}}
    \hline
        \textbf{(ao)} Skewers with charred meat, roasted golden potatoes, and roasted red tomatoes served on a white plate with a brown rim. The background is softly blurred with table elements visible. \\ \hline
        \textbf{(ap)} Blossom fruit. beautiful spring. \\ \hline
        \textbf{(aq)} An adult cow and calf lying on straw inside a wooden barn. The cow is brown with white patches, the calf mostly white with brown markings, lit by soft natural light. \\ \hline
        \textbf{(ar)} Slow Down Tours on the train. \\ \hline
        \textbf{(as)} A calm beach scene with a large wet boulder in the foreground, sandy shore, gentle blue waves, and a clear sky creating a tranquil atmosphere. \\ \hline
        \textbf{(at)} A forest landscape with green, red, and yellow foliage in the foreground and a mountain covered in autumn leaves in the background. \\ \hline
        \textbf{(au)} Gorgeous portrait of a blue peacock with silky blue feathers. \\ \hline
        \textbf{(av)} A large rock formation covered in green plants with light blue water in front and a cloudy sky above. \\ \hline
        \textbf{(aw)} A tall glass of green beverage with a purple lid on a wooden surface, surrounded by colorful signage with Chinese characters, suggesting a café or food stall setting. \\ \hline
        \textbf{(ax)} An ancient stone wall with Mayan carvings depicting figures, animals, and geometric shapes, weathered and textured in light gray stone. \\ \hline
        \textbf{(ay)} Abandoned barn in Sauk County, Wisconsin - LOC's Public Domain Archive Public Domain Search. \\ \hline
        \textbf{(az)} Wrapped Up In Wool Penguin. \\ \hline
        \textbf{(ba)} A utility pole with multiple electrical wires and a street lamp in the foreground, with a dark wooden building and partly cloudy blue sky behind it. \\ \hline
        \textbf{(bb)} A cozy bedroom with a large bed, white comforter, beige walls, framed pictures, seating furniture, a rug, and a vintage chandelier with candle-style lights. \\ \hline
        \textbf{(bc)} A golden porcelain Turkish coffee cup and saucer with ornate detailing and a lion-shaped handle, isolated on a reflective surface against a white background. \\ \hline
        \textbf{(bd)} A person in a white dress holding a bouquet of white and pink roses with green foliage, set against a natural green background. \\ \hline
        \textbf{(be)} A detailed Carcassonne Mini board game tile depicting a medieval village with church, temple, river, and trees, rendered in a realistic illustrated style. \\ \hline
        \textbf{(bf)} A carved Halloween pumpkin with triangular eyes and jagged mouth, wearing a black witch hat, resting on dark leaves in a spooky setting. \\ \hline
        \textbf{(bg)} A white horse standing inside a fenced area with a red-and-white striped barrier, facing the camera with alert ears and trees in the background. \\ \hline
        \textbf{(bh)} A collage of contemporary furniture pieces including beds, shelving, dressers, mirrors, and chandeliers displayed across modern bedroom interiors. \\ \hline
    \end{tabular}
    \caption{Prompts used to generate images in \autoref{fig:extended_image_generations_3}.}
    \label{tab:generated_images_prompts_3}
\end{table*}

\begin{table*}[t]
    \centering
    \renewcommand{\arraystretch}{1}
    \begin{tabular}{p{0.98\textwidth}}
    \hline
        \textbf{(bi)} A wide shot of a valley surrounded by snow-covered mountains under a cloudy sky, evenly lit by natural daylight. \\ \hline
        \textbf{(bj)} A white lighthouse on a rocky cliff surrounded by green vegetation, overlooking the ocean beneath a clear blue sky with scattered clouds. \\ \hline
        \textbf{(bk)} A creeping thistle flower with a vibrant pink center on a sepia-toned old paper background, softly lit with blank space for text. \\ \hline
        \textbf{(bl)} A dimly lit dining table set with plates of food and wine glasses. Three wine bottles are visible, chairs surround the table, and a wooden wall adds a cozy mood. \\ \hline
        \textbf{(bm)} Streets of Kanazawa - japanese, japan, kanazawa, street, building, oriental, lantern. \\ \hline
        \textbf{(bn)} A white bowl of soup with meat, carrots, and greens on a wooden table, with salt and pepper shakers, a tomato, and a beige napkin nearby. \\ \hline
        \textbf{(bo)} A close-up view of stalactites hanging from a cave ceiling, with rugged textured walls and contrasting light and shadow highlighting natural formations. \\ \hline
        \textbf{(bp)} A rocky shoreline at Nha Trang beach, Vietnam, with turquoise water creating white foam, calm sea beyond, and a clear sky with scattered clouds. \\ \hline
        \textbf{(bq)} A spider positioned at the center of its web, facing the camera, surrounded by green leaves with a softly blurred background and light specks. \\ \hline
        \textbf{(br)} Ship wreck "Superior Producer" in turquoise water of coral reef in Caribbean Sea. \\ \hline
        \textbf{(bs)} A bright cityscape with numerous beige buildings of varying roof styles, trees on the left, and clear sunny weather. \\ \hline
        \textbf{(bt)} A tree with a thick trunk near a dirt path leading into a canyon filled with orange rock formations and scattered pine trees. \\ \hline
        \textbf{(bu)} A cappuccino in a white porcelain cup and saucer on a wooden table, topped with a floral chocolate design and photographed with a soft blur. \\ \hline
        \textbf{(bv)} A quiet winter forest with a snow-covered path winding through densely packed trees under an overcast sky. \\ \hline
        \textbf{(bw)} A four-image collage showing stone buildings, a glass structure with purple light, and buildings near large bodies of water. \\ \hline
        \textbf{(bx)} A collage of 36 cat photos arranged in a uniform grid, each cat looking at the camera against a bright background. \\ \hline
        \textbf{(by)} A lush hillside with green grass, trees, and a large boulder in the foreground, with cloudy skies over the Bulbul hills in the background. \\ \hline
        \textbf{(bz)} A panoramic cityscape showing colorful residential buildings in the foreground and an industrial area with smokestacks under a clear blue sky. \\ \hline
        \textbf{(ca)} A vibrant garden with a tree full of ripe oranges near a swimming pool, featuring potted plants and an umbrella in bright light. \\ \hline
        \textbf{(cb)} Different spring blossoms in a little bottle with nature background. \\ \hline
    \end{tabular}
    \caption{Prompts used to generate images in \autoref{fig:extended_image_generations_4}.}
    \label{tab:generated_images_prompts_4}
\end{table*}

\clearpage
\newpage
\section{Contributions}
\label{sec:contributions}

All authors contributed to writing this paper, designing the experiments and discussing results at each stage of the project.

\paragraph{Code.} General training code was written by Jason Ramapuram, Victor Turrisi, Louis Béthune and Vishnu Banna. Bruno Mlodozeniec and Dan Busbridge extended the baseline functional optimizers to support MuP and CompleteP. Evaluators were written by Pau Rodriguez Lopez in collaboration with Louis Béthune and Arno Blaas.  

\paragraph{Experiments.} Scaling laws experiments were sculpted and executed by Louis Béthune in discussions with Amitis Shidani, Pierre Ablin and Jason Ramapuram. Main model (\Cref{sec:architecture}) was trained by Victor Turrisi in discussions with Jason Ramapuram and Louis Béthune. Bruno Kacper Mlodozeniec wrote and executed the per-module hyper-parameter search (\Cref{app:per_module_hyperparam}) and crafted $B_\textbf{crit}$ experimental procedure that was executed by Louis Béthune (\Cref{sec:b_opt}). Inference ablations were crafted and executed by Lokesh Boominathan and Nikhil Bhendawade in discussions with Theo X. Olausson (\Cref{sec:inference_ablations}). Data mixtures experiments (\Cref{sec:modality_mixing}) were designed by Pierre Ablin and Louis Béthune, and executed by Nikhil Bhendawade and Louis Béthune. João Monteiro crafted and executed the anti-masking experiments (\Cref{sec:anti_masking_ablation}) in discussions with Victor Turrisi, Jason Ramapuram, Louis Béthune and Amitis Shidani. Tokenizers were trained and benchmarked by Paul Dixon (\Cref{app:audio_tokenizer_ablations}) and Devon Hjelm (\Cref{app:image_tokenizer_ablations}) in discussions with Jason Ramapuram and Victor Turrisi. 

\paragraph{Data.} The data loading library was built by Victor Turrisi in collaboration with Louis Béthune. Data collection and pre-processing was done by Louis Béthune, Victor Turrisi, Joris Pelemans, Kari Noriy, Jason Ramapuram, Luca Zappella and Nikhil Bhendawade.

\paragraph{General Infrastructure.} Nick Henderson built the pipeline to build docker containers with all optimizations for networking and high-performance training.

\paragraph{Theoretical formulation and situating work.} The theoretical framework (\Cref{sec:method,app:weighting}) was crafted by Amitis Shidani in discussions with Pierre Ablin, Devon Hjelm and Arno Blaas. Grounding work with respect to relevant literature executed by Arno Blaas (\Cref{sec:background}) in discussions with Amitis Shidani and Jason Ramapuram.

\paragraph{Project Organization and Tech Lead.} Overall project organization and guidance enabled by Irina Belousova, Luca Zappella, Russ Webb and Jason Ramapuram. Jason Ramapuram organized, setup scientific objectives, provided technical leadership and setup the preliminary fault tolerant, distributed scalable code-base for the project.

\section{Acknowledgments}
\label{sec:acknowledgements}

We thank
Samy Bengio,
Jerremy Holland,
Erik Wijmans, 
David Koski, 
Miguel Sarabia del Castillo,
for their helpful feedback and critical discussions throughout the process of writing this paper;
Michael Brooks,
Denise Hui,
Li Li, 
Rajat Phull,
Evan Samanas, 
Guillaume Seguin, 
and the wider Apple infrastructure team for assistance with developing and running scalable, fault tolerant code. 
Names are in alphabetical order by last name within group.

\stopcontents[sections]

\end{document}